\documentclass[10pt,twocolumn,letterpaper]{article}

\usepackage{cvpr}
\usepackage{times}
\usepackage{epsfig}
\usepackage{graphicx}
\usepackage{amsmath}
\usepackage{amssymb}
\usepackage{subfigure}
\usepackage{multirow}
\usepackage{color}

\usepackage[pagebackref=true,breaklinks=true,letterpaper=true,colorlinks,bookmarks=false]{hyperref}

\cvprfinalcopy 

\def\cvprPaperID{1315} 

\ifcvprfinal\fi

\begin{document}

\title{Fast and Accurate Single Image Super-Resolution via Information Distillation Network}

\author{Zheng~Hui, Xiumei~Wang, Xinbo~Gao\\
School of Electronic Engineering, Xidian University\\
Xi'an, China\\
{\tt\small zheng\_hui@aliyun.com, wangxm@xidian.edu.cn, xbgao@mail.xidian.edu.cn}
}

\maketitle
\thispagestyle{empty}

\begin{abstract}
   Recently, deep convolutional neural networks (CNNs) have been demonstrated remarkable progress on single image super-resolution. However, as the depth and width of the networks increase, CNN-based super-resolution methods have been faced with the challenges of computational complexity and memory consumption in practice. In order to solve the above questions, we propose a deep but compact convolutional network to directly reconstruct the high resolution image from the original low resolution image. In general, the proposed model consists of three parts, which are feature extraction block, stacked information distillation blocks and reconstruction block respectively. By combining an enhancement unit with a compression unit into a distillation block, the local long and short-path features can be effectively extracted. Specifically, the proposed enhancement unit mixes together two different types of features and the compression unit distills more useful information for the sequential blocks. In addition, the proposed network has the advantage of fast execution due to the comparatively few numbers of filters per layer and the use of group convolution. Experimental results demonstrate that the proposed method is superior to the state-of-the-art methods, especially in terms of time performance. Code is available at \url{https://github.com/Zheng222/IDN-Caffe}.
\end{abstract}

\section{Introduction}
Single image super-resolution (SISR) is a classical problem in low-level computer vision, which reconstructs a high-resolution (HR) image from a low-resolution (LR) image. Actually, an infinite number of HR images can get the same LR image by downsampling. Hence, the SR problem is inherently ill-posed and no unique solution exists.  In order to mitigate this problem, numerous SISR methods have been proposed in the literature, including interpolation-based methods, reconstruction-based methods and example-based methods. Since the former two kinds of methods usually suffer dramatically drop in restoration performance with larger upscaling factors, the recent SR methods fall into the example-based methods which try to learn prior knowledge from LR and HR pairs.

\begin{figure}[!t]
	\begin{center}
		\includegraphics[width=0.48\textwidth]{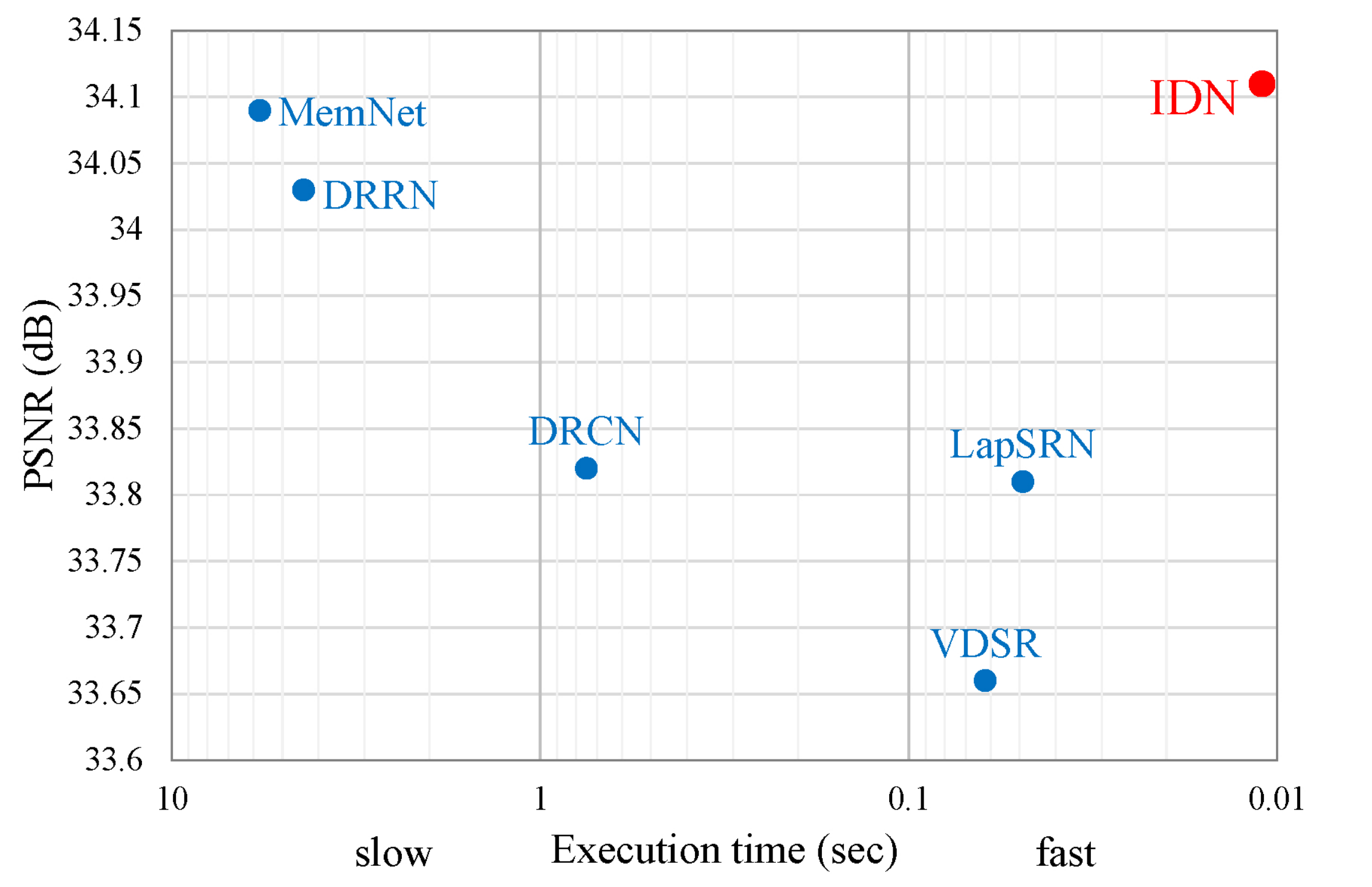}
	\end{center}
	\caption{Speed and accuracy trade-off. The average PSNR and the average inference time for upscaling $3 \times$ on Set5. The IDN is faster than other methods and achieves the best performance at the same time.}
	\label{fig:execution_time}
\end{figure}

Recently, due to the strength of deep convolutional neural network (CNN), many CNN-based SR methods try to train a deep network to gain better reconstruction performance. Kim~\etal propose a 20-layer CNN model known as VDSR~\cite{VDSR}, which adopts residual learning and adaptive gradient clipping to ease training difficulty. To control the model parameters, the authors construct a deeply-recursive convolutional network (DRCN)~\cite{DRCN} by adopting recursive layer. To mitigate training difficulty, Mao~\etal propose a very deep residual encoder-decoder network (RED)~\cite{RED}, which consists of a series of convolutional and subsequent transposed convolution layers to learn end-to-end mappings from the LR images to the ground truths. Tai~\etal propose a deep recursive residual network (DRRN)~\cite{DRRN}, which employs parameters sharing strategy to alleviate the requirement of enormous parameters of the very deep networks.

Although achieving prominent performance, most of deep networks still have some drawbacks. Firstly, in order to achieve better performance, deepening or widening the network has been a design trend. But the result is that these methods demand large computational cost and memory consumption, which are less applicable in practice, such as mobile and embedded vision applications. Moreover, the traditional convolutional networks usually adopt cascaded network topologies,~\eg, VDSR~\cite{VDSR} and DRCN~\cite{DRCN}. In this way, the feature maps of each layer are sent to the sequential layer without distinction. However, Hu~\etal~\cite{SENet} experimentally demonstrate that adaptively recalibrating channel-wise features responses can improve the representational power of a network.

To address these drawbacks, we propose a novel information distillation network (IDN) with lightweight parameters and computational complexity as illustrated in Figure~\ref{fig:structure}. In the proposed IDN, a feature extraction block (FBlock) first extracts features from the LR image. Then, multiple information distillation blocks (DBlocks) are stacked to progressively distill residual information. Finally, a reconstruction Block (RBlock) aggregates the obtained HR residual representations to generate the residual image. To get a HR image, we implement an element-wise addition operation on the residual image and the upsampled LR image.

The key component of IDN is the information distillation block, which contains an enhancement unit and a compression unit. The enhancement unit mainly comprises two shallow convolutional networks as illustrated in Figure~\ref{fig:Eunit}. Each of them is a three-layer shallow module. The feature maps of the first module are extracted through a short path (3-layer). Thus, they can be regarded as the local short-path features. Considering that the deep networks have more expressive power, we send a portion of the local short-path features to another module. By this way, the feature maps of the second module naturally become the local long-path features. Different from the approach in~\cite{SENet}, we divide feature maps into two parts. One part represents reserved short-path features and another expresses the short-path features that will be enhanced. After getting long and short-path feature maps, we aggregate these two types of features for gaining more abundant and efficient information. In summary, the enhancement unit is mainly to improve the representation power of the network. As for the compression unit, we adopt a simple convolutional layer to compress the redundancy information in features of the enhancement unit.

The main contributions of this work are summarized as follows:
\begin{itemize}
	\item The proposed IDN extracts feature maps directly from LR images and employs multiple cascaded DBlocks to generate the residual representations in HR space. In each DBlock, the enhancement unit gathers more information as much as possible and the compression unit distills more useful information. As a result, IDN achieves competitive results in spite of using less number of convolutional layer.
	\item Due to the concise structure of the proposed  IDN, it is much faster than several CNN-based SR methods,~\eg, VDSR~\cite{VDSR}, DRCN~\cite{DRCN}, LapSRN~\cite{LapSRN}, DRRN~\cite{DRRN} and MemNet~\cite{MemNet} as illustrated in Figure~\ref{fig:execution_time}. Only the proposed method achieves real-time speed and maintains better reconstruction accuracy.
\end{itemize}

\section{Related Work}

Single image super-resolution has been extensively studied in these years. In this section, we will focus on recent example-based and neural network based approaches.

\subsection{Self-example based methods}

The self-example based methods exploit the self-similarity property and extract example pairs merely from the LR image across different scales. This type of methods usually works well in the images containing repetitive patterns and textures but lacks the richness of image structures outside the input image and thus fails to generate satisfactory prediction for images of other classes. Huang \etal~\cite{Urban100} extend self-similarity based SR to handle the affine and perspective deformation.

\subsection{External-example based methods}

The external-example based methods learn a mapping between LR and HR patches from external datasets. This type of approaches usually focuses on how to learn a compact dictionary or manifold space to relate LR/HR patches, such as nearest neighbor~\cite{NN}, manifold embedding~\cite{manifold_embedding}, random forest~\cite{random_forest} and sparse representation~\cite{sparse_coding,sparse_coding_TIP}. While these approaches are effective, the extracted features and mapping functions are not adaptive, which may not be optimal for generating high-quality SR images.

\begin{figure*}[htb]
	\begin{center}
		\includegraphics[width=0.9\textwidth]{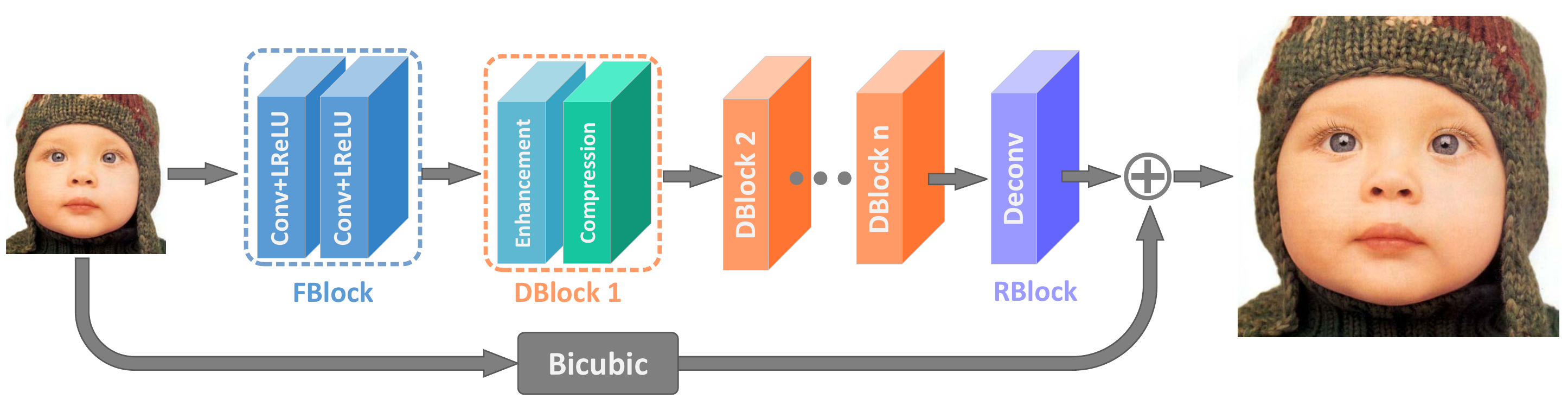}
	\end{center}
	\caption{Architecture of the proposed network.}
	\label{fig:structure}
\end{figure*}

\subsection{Convolutional neural networks based methods}
Recently, inspired by the achievement of many computer vision tasks tackled with deep learning, neural networks have been achieved dramatic improvement in SR. Dong~\etal~\cite{SRCNN,SRCNN-Ex} first exploit a three-layer convolutional neural network, named SRCNN, to jointly optimize the feature extraction, non-linear mapping and image reconstruction stages in an end-to-end manner. Afterwards Shi~\etal~\cite{ESPCN} propose an efficient sub-pixel convolutional neural network (ESPCN), which extracts feature maps in the LR space and replaces the bicubic upsampling operation with an efficient sub-pixel convolution. Dong~\etal~\cite{FSRCNN} adopt deconvolution to accelerate SRCNN in combination with smaller filter sizes and more convolution layers. Kim~\etal~\cite{VDSR} propose a very deep CNN model with global residual architecture to achieve superior performance, which utilizes contextual information over large image regions. Another network designed by Kim~\etal~\cite{DRCN}, which has recursive convolution with skip connection to avoid introducing additional parameters when the depth is increasing. Mao~\etal~\cite{RED} tackle the general image restoration problem with encoder-decoder networks and symmetric skip connections. Lai \etal~\cite{LapSRN} propose the laplacian pyramid super-resolution network (LapSRN) to address the speed and accuracy of SR problem, which takes the original LR images as input and progressively reconstructs the sub-band residuals of HR images. Tai~\etal~\cite{DRRN} propose the deep recursive residual network to effectively build a very deep network structure for SR, which weighs the model parameters against the accuracy. The authors also present a very deep end-to-end persistent memory network (MemNet)~\cite{MemNet} for image restoration task, which tackles the long-term dependency problem in the previous CNN architectures. Sajjadi~\etal~\cite{EnhanceNet} propose a novel combination of automated texture synthesis with a perceptual loss focusing on creating realistic textures at a high magnification ratio of 4.

\section{Proposed Method}
In this section, we first describe the proposed model architecture and then suggest the enhancement unit and the compression unit, which are the core of the proposed method.

\subsection{Network structure}

The proposed IDN, as shown in Figure~\ref{fig:structure}, consists of three parts: a feature extraction block (FBlock), multiple stacked information distillation blocks (DBlocks) and a reconstruction block (RBlock). Here, we denote $x$ and $y$ as the input and the output of IDN. With respect to FBlock, two $3\times3$ convolutional layers are utilized to extract the feature maps from the original LR image. This procedure can be expressed as
\begin{equation}
{B_0} = f\left( x \right),
\end{equation}
where $f$ represents the feature extraction function and ${B_0}$ denotes the extracted features and servers as the input to the following stage. The next part is composed of multiple information distillation blocks by using chained mode. Each block contains an enhancement unit and a compression unit with stacked style. This process can be formulated as
\begin{equation}
{B_k} = {F_k}\left( {{B_{k - 1}}} \right),k = 1, \cdots ,n,
\end{equation}
where ${F_k}$ denotes the $k$-th DBlock function, ${{B_{k - 1}}}$ and ${B_k}$ indicate the input and output of the $k$-th DBlock respectively. Finally, we take a transposed convolution without activation function as the RBlock. Hence, the IDN can be expressed as
\begin{equation}
y = R\left( {{F_n}\left( {{B_{n - 1}}} \right)} \right) + U\left( x \right),
\end{equation}
where $R$, $U$ denote the RBlock and bicubic interpolation operation respectively.
\subsubsection{Loss function}
We consider two loss functions that measure the difference between the predicted HR image $\hat I$ and the corresponding ground-truth $I$. The first one is mean square error (MSE), which is the most widely used loss function for general image restoration as defined below:
\begin{equation}
{l_{MSE}} = \frac{1}{N}\sum\limits_{i = 1}^N {\left\| {{I_i} - {{\hat I}_i}} \right\|_2^2}.
\end{equation}
However, Lim \etal~\cite{EDSR} experimentally demonstrate that training with MSE loss is not a good choice. The second loss function is mean absolute error (MAE), which is formulated as follows:
\begin{equation}
{l_{MAE}} = \frac{1}{N}\sum\limits_{i = 1}^N {{{\left\| {{I_i} - {{\hat I}_i}} \right\|}_1}}.
\end{equation}
We empirically found that our model with MSE loss can improve performance of a trained network with MAE loss. Therefore, we first train the network with MAE loss and then fine-tune it by MSE loss.

\begin{figure}[htb]
	\begin{center}
		\includegraphics[width=0.35\linewidth]{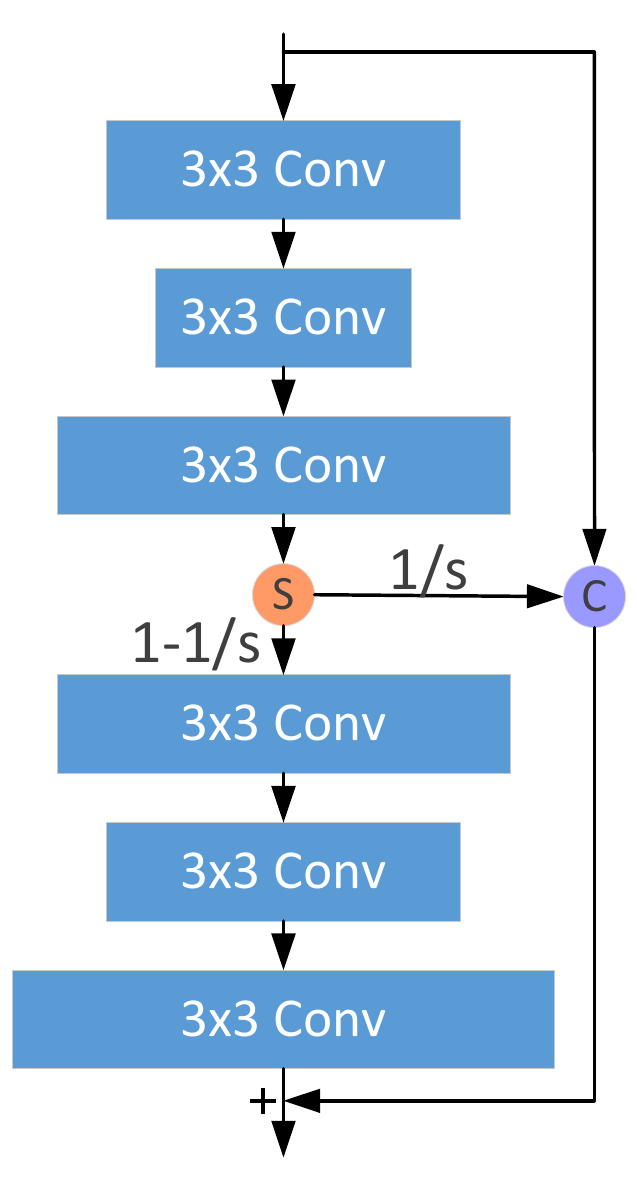}
	\end{center}
	\caption{The architecture of enhancement unit in the proposed model. Orange circle represents slice operation and purple circle indicates concatenation operation in channel dimension.}
	\label{fig:Eunit}
\end{figure}

\subsection{Enhancement unit}
As shown in Figure~\ref{fig:Eunit}, enhancement unit can be roughly divided into two modules, one is the above three convolutions and another is the below three convolutions. The above module has three $3 \times3 $ convolutions, each of them is followed by a leaky rectified linear unit (LReLU) activation function, which is omitted here. Let's denote the feature map dimensions of the $i$-th layer as ${D_i}\left( {i = 1, \cdots ,6} \right)$. In that way, the relationship of the convolutional layers can be expressed as
\begin{equation}
{D_3} - {D_1} = {D_1} - {D_2} = d,
\end{equation}
where $d$ denotes the difference between the first layer and the second layer or between the first layer and the third layer. Similarly, the dimension of channels in the below module also has this relation and can be described as follows:
\begin{equation}
{D_6} - {D_4} = {D_4} - {D_5} = d,
\end{equation}
where ${D_4} = {D_3}$. The above module is composed of three cascaded convolution layers with LReLUs, and the output of the third convolution layer is sliced into two segments. Supposing the input of this module is ${B_{k - 1}}$, we have
\begin{equation}
P_1^k = {C_a}\left( {{B_{k - 1}}} \right),
\end{equation}
where ${{B_{k - 1}}}$ denotes the output of previous block and meanwhile is the input of present block, ${C_a}$ indicates chained convolutions operation and $P_1^k$ is the output of the above module in the $k$-th enhancement unit. The feature maps with $ \frac{{{D_3}}}{s}$ dimensions of $P_1^k$ and the input of the first convolutional layer are concatenated in the channel dimension,
\begin{equation}
{R^k} = C\left( {S\left( {P_1^k,1/s} \right),{B_{k - 1}}} \right),
\end{equation}
where $C$, $S$ represent concatenation operation and slice operation respectively. Specifically, we know the dimension of $P_1^k$ is ${D_{\rm{3}}}$. Therefore, $S\left( {P_1^k,1/s} \right)$ denotes that $\frac{{{D_3}}}{s}$ dimensions features are fetched from ${P_1^k}$. Moreover, $S\left( {P_1^k,1/s} \right)$ concatenates features with ${B_{k - 1}}$ in channel dimension. The purpose is to combine the previous information with some current information. It can be regarded as partially retained local short-path information. We take the rest of local short-path information as the input of the below module, which mainly further extracts long-path feature maps,
\begin{equation}
P_2^k = {C_b}\left( {S\left( {P_1^k,1 - 1/s} \right)} \right),
\end{equation}
where $P_2^k$, ${C_b}$ are the output and stacked convolution operations of the below module respectively. Finally, as shown in Figure~\ref{fig:Eunit}, the input information, the reserved local short-path information and the local long-path information are aggregated. Therefore, the enhancement unit can be formulated as
\begin{equation}
\begin{array}{l}
{P^k} = P_2^k + {R^k} = {C_b}\left( {S\left( {{C_a}\left( {{B_{k - 1}}} \right),1 - 1/s} \right)} \right)\\
+ C\left( {S\left( {{C_a}\left( {{B_{k - 1}}} \right),1/s} \right),{B_{k - 1}}} \right),
\end{array}
\end{equation}
where ${P^k}$ is the output of enhancement unit. At this point, local long-path features $P_2^k$ and the combination of local short-path features and the untreated features ${R^k}$ are utilized without exception by a compression unit.

\subsection{Compression unit}
We achieve compression mechanism by taking advantage of a $1\times1$ convolution layer. Concretely, the outputs of the enhancement unit are sent to a $1\times1$ convolution layer, which acts as dimensionality reduction or distilling relevant information for the later network. Thus, the compression unit can be formulated as
\begin{equation}
{B_k} = f_F^k\left( {{P^k}} \right) = \alpha _F^k\left( {W_F^k\left( {{P^k}} \right)} \right),
\end{equation}
where $f_F^k$ denotes the function of the $1\times1$ convolution layer ($\alpha _F^k$ denotes the activation function and $W_F^k$ is the weight parameters).

\section{Experiments}
\subsection{Datasets}
\subsubsection{Training datasets}
By following~\cite{VDSR,LapSRN,DRRN,MemNet}, we use 91 images from Yang~\etal~\cite{sparse_coding_TIP} and 200 images from Berkeley Segmentation Dataset (BSD)~\cite{BSD} as the training data. As in~\cite{DRRN}, to make full use of the training data, we apply data augmentation in three ways: (1) Rotate the images with the degree of ${90^ \circ }$, ${180^ \circ }$ and ${270^ \circ }$. (2) Flip images horizontally. (3) Downscale the images with the factor of $0.9$, $0.8$, $0.7$ and $0.6$.

\subsubsection{Testing datasets}
The proposed method is evaluated on four widely used benchmark datasets: Set5~\cite{Set5}, Set14~\cite{Set14}, BSD100~\cite{BSD}, Urban100~\cite{Urban100}. Among these datasets, Set5, Set14 and BSD100 consist of natural scenes and Urban100 contains challenging urban scenes images with details in different frequency bands. The ground truth images are downscaled by bicubic interpolation to generate LR/HR image pairs for both training and testing datasets. We convert each color image into the YCbCr color space and only process the Y-channel, while color components are simply enlarged using bicubic interpolation.
\begin{table}[htb]
	\small
	\centering
	\begin{tabular}{|c|c|c|}
		\hline
		Scale & Training & Fine-tuning \\
		\hline\hline
		2 & ${29^2}/{57^2}$ & ${39^2}/{77^2}$ \\
		3 & ${15^2}/{43^2}$ & ${26^2}/{76^2}$ \\
		4 & ${11^2}/{41^2}$ & ${19^2}/{73^2}$ \\
		\hline
	\end{tabular}
	\caption{The sizes of training and fine-tuning sub-images for different scaling factors.}
	\label{samples_setting}
\end{table}

\subsection{Implementation details}
For preparing the training samples, we first downsample the original HR images with upscaling factor $m\left( {m = 2,3,4} \right)$ by using the bicubic interpolation to generate the corresponding LR images and then crop the LR training images into a set of $ {l_{sub}} \times {l_{sub}}$ size sub-images. The corresponding HR training images are divided into $m{l_{sub}} \times m{l_{sub}}$ size sub-images. As the proposed model is trained using the \emph{Caffe} package~\cite{Caffe}, its transposed convolution filters will generate the output with size ${\left( {m{l_{sub}} - m + 1} \right)^2}$ instead of ${\left( {m{l_{sub}}} \right)^2}$. So we should crop $\left( {m - 1} \right)$-pixel borders on the HR sub-images. Since the minimum size picture ``t20'' in the 291 dataset is a $78 \times 78$ size image, the maximum size of the sub-image we can crop on the LR image is $26 \times 26$ for maintaining data integrity when scaling factor $m=3$. However, the training process will be unstable due to the larger size training samples equipped with the larger learning rate by using \emph{Caffe} package. Therefore, ${15^2}/{43^2}$ training pairs are generated for training stage and ${26^2}/{76^2}$ LR/HR sub-images pairs are utilized for fine-tuning phase. The learning rate is initially set to $1e-4$ and decreases by the factor of $10$ during fine-tuning phase. In this way, the sizes of training and fine-tuning samples are shown in Table~\ref{samples_setting}.

Taking into account the trade-off between the execution time and the reconstruction performance, we construct a 31-layer network that denoted as IDN. This model has 4 DBlocks, and the parameters ${D_3}$, $d$ and $s$ of enhancement unit in each block are set to 64, 16 and 4 respectively. To reduce the parameters of network, we use the grouped convolution layer~\cite{Xception,ResNeXt} in the second and fourth layers in each enhancement unit with 4 groups. In addition, the transposed convolution adopts $17 \times17$ filters for all scaling factors and the negative scope of LReLU is set as 0.05. We initialize the weights by using the method proposed in~\cite{Msra} and the biases are set to zero. The proposed network is optimized using Adam~\cite{Adam}. We set the parameters of mini-batch size and weight decay to 64 and $1e-4$ respectively. In order to get better initialization parameters, we empirically pre-train the proposed model with ${10^5}$ iterations and take these parameters as the initial values of the IDN. Training a IDN roughly takes a day with a TITAN X GPU on the $2 \times$ model.

\begin{figure}[htb]
	\centering
	\subfigure[residual image]{\label{residual_image}
		\includegraphics[width=0.2\textwidth]{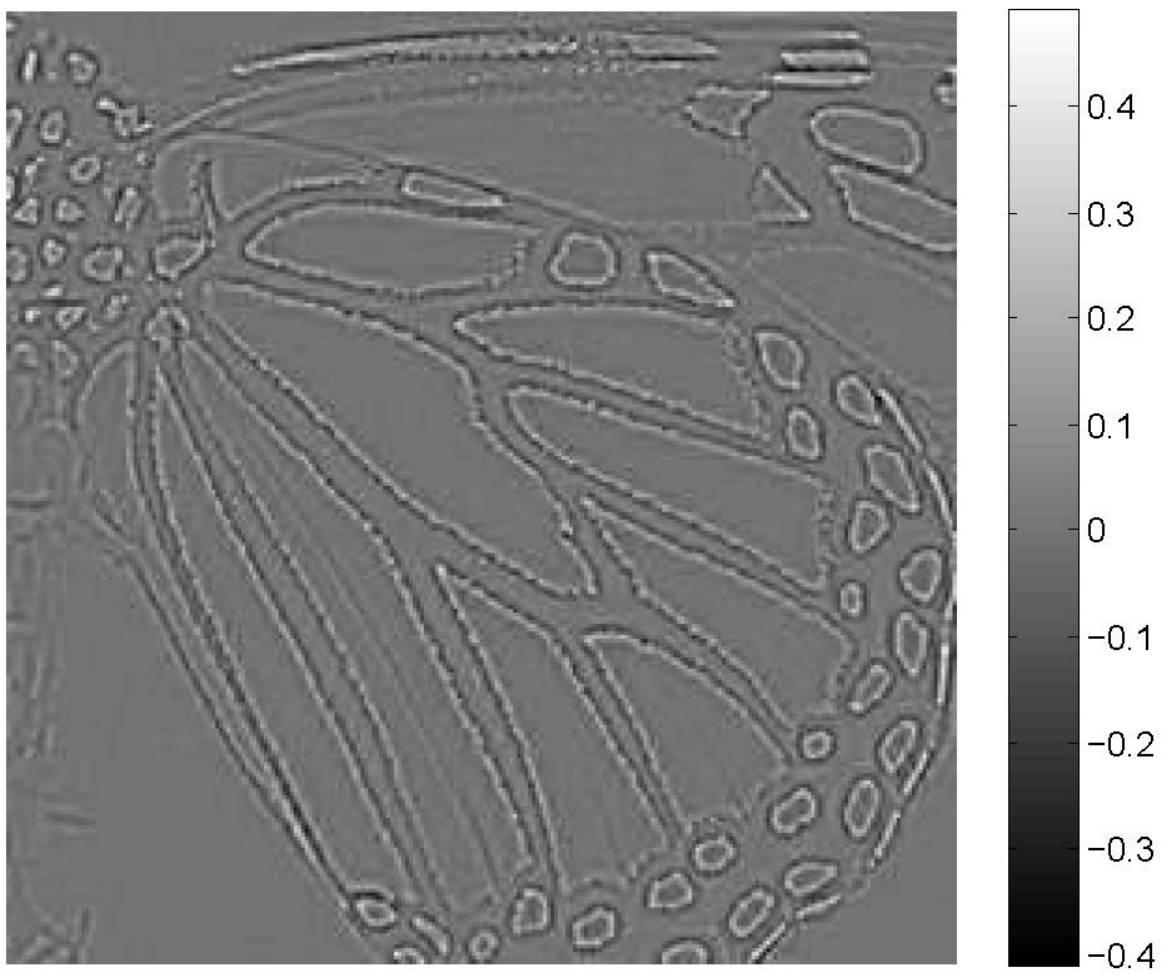}}
	\hfil
	\subfigure[data distribution histogram]{\label{data_distribution}
		\includegraphics[width=0.2\textwidth]{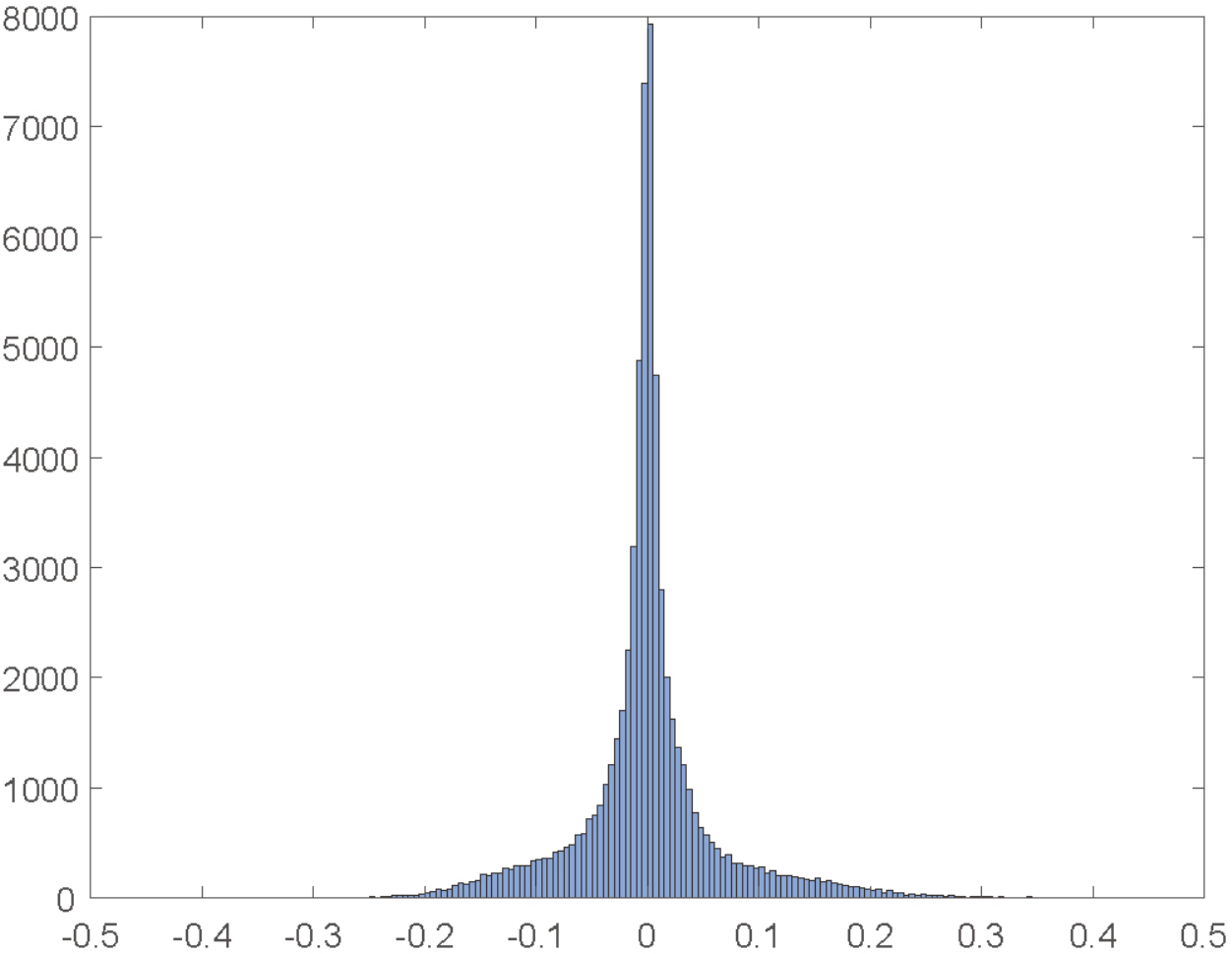}}
	\caption{The residual image and its data distribution of the ``butterfly'' image from Set5 dataset.}
	\label{fig:real_image}
\end{figure}

\begin{figure}[htb]
	\centering
	\subfigure[the average feature maps of enhancement units]{\includegraphics[width=0.48\textwidth]{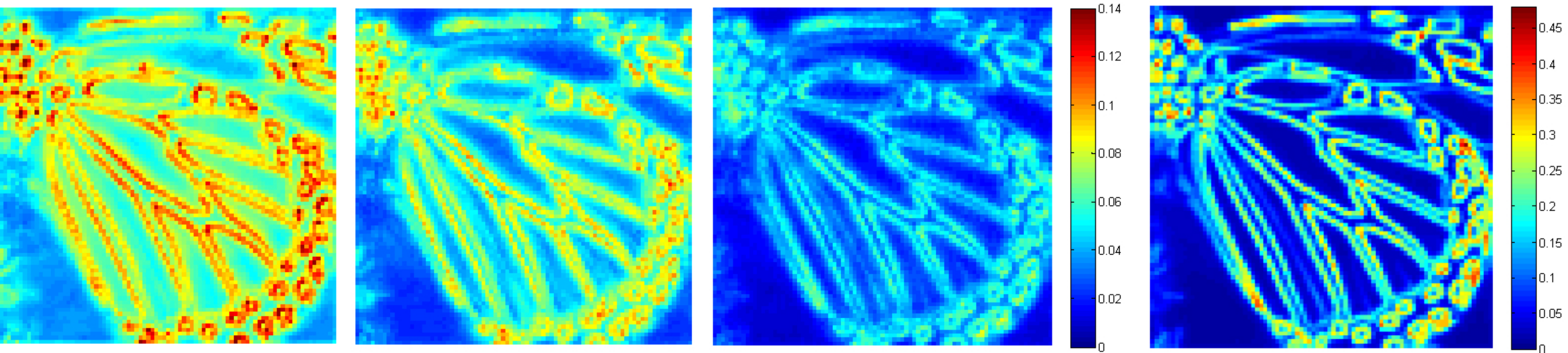}%
		\label{Eunit_analyse}}
	\hfil
	\subfigure[the average feature maps of compression units]{\includegraphics[width=0.48\textwidth]{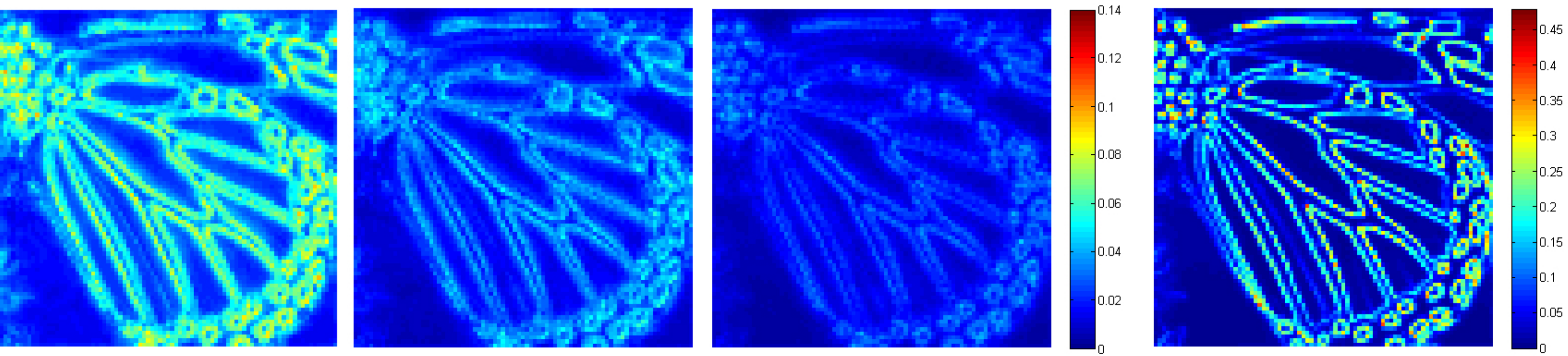}%
		\label{CUnit_analyse}}
	\caption{Visualization of the average feature maps.}
	\label{network_analyse}
\end{figure}

\begin{table*}[htp]
	\small 
	\begin{center}
		\begin{tabular}{|c|c|c|c|c|c|c|c||c|}
			\hline
			Dataset & Scale & Bicubic & VDSR~\cite{VDSR} &¡¡DRCN~\cite{DRCN} & LapSRN~\cite{LapSRN} & DRRN~\cite{DRRN} & MemNet~\cite{MemNet} & IDN~(Ours) \\
			\hline
			\hline
			\multirow{3}{*}{Set5} & $\times 2$ & 33.66/0.9299 & 37.53/0.9587 & 37.63/0.9588 &¡¡37.52/0.9591 & 37.74/0.9591 & \textcolor[rgb]{0.00,0.07,1.00}{37.78}/\textcolor[rgb]{0.00,0.07,1.00}{0.9597} & \textcolor[rgb]{1.00,0.00,0.00}{37.83}/\textcolor[rgb]{1.00,0.00,0.00}{0.9600} \\
			
			& $\times 3$ & 30.39/0.8682 & 33.66/0.9213 & 33.82/0.9226 & 33.81/0.9220 & 34.03/0.9244 & \textcolor[rgb]{0.00,0.07,1.00}{34.09}/\textcolor[rgb]{0.00,0.07,1.00}{0.9248} & \textcolor[rgb]{1.00,0.00,0.00}{34.11}/\textcolor[rgb]{1.00,0.00,0.00}{0.9253} \\
			
			& $\times 4$ & 28.42/0.8104 & 31.35/0.8838 & 31.53/0.8854 & 31.54/0.8852 & 31.68/0.8888 & \textcolor[rgb]{0.00,0.07,1.00}{31.74}/\textcolor[rgb]{0.00,0.07,1.00}{0.8893} & \textcolor[rgb]{1.00,0.00,0.00}{31.82}/\textcolor[rgb]{1.00,0.00,0.00}{0.8903} \\
			\hline
			\hline
			\multirow{3}{*}{Set14} & $\times 2$ & 30.24/0.8688 & 33.03/0.9124 & 33.04/0.9118 & 32.99/0.9124 & 33.23/0.9136 & \textcolor[rgb]{0.00,0.07,1.00}{33.28}/\textcolor[rgb]{0.00,0.07,1.00}{0.9142} & \textcolor[rgb]{1.00,0.00,0.00}{33.30}/\textcolor[rgb]{1.00,0.00,0.00}{0.9148} \\
			
			& $\times 3$ & 27.55/0.7742 & 29.77/0.8314 & 29.76/0.8311 & 29.79/0.8325 & 29.96/0.8349 & \textcolor[rgb]{1.00,0.00,0.00}{30.00}/\textcolor[rgb]{0.00,0.07,1.00}{0.8350} & \textcolor[rgb]{0.00,0.07,1.00}{29.99}/\textcolor[rgb]{1.00,0.00,0.00}{0.8354} \\
			
			& $\times 4$ & 26.00/0.7027 & 28.01/0.7674 & 28.02/0.7670 & 28.09/0.7700 & 28.21/0.7721 & \textcolor[rgb]{1.00,0.00,0.00}{28.26}/\textcolor[rgb]{0.00,0.07,1.00}{0.7723} & \textcolor[rgb]{0.00,0.07,1.00}{28.25}/\textcolor[rgb]{1.00,0.00,0.00}{0.7730} \\
			
			\hline
			\hline
			\multirow{3}{*}{BSD100} & $\times 2$ & 29.56/0.8431 & 31.90/0.8960 & 31.85/0.8942 & 31.80/0.8952 & \textcolor[rgb]{0.00,0.07,1.00}{32.05}/0.8973 & \textcolor[rgb]{1.00,0.00,0.00}{32.08}/\textcolor[rgb]{0.00,0.07,1.00}{0.8978} & \textcolor[rgb]{1.00,0.00,0.00}{32.08}/\textcolor[rgb]{1.00,0.00,0.00}{0.8985} \\
			
			& $\times 3$ & 27.21/0.7385 & 28.82/0.7976 & 28.80/0.7963 & 28.82/0.7980 & \textcolor[rgb]{0.00,0.07,1.00}{28.95}/\textcolor[rgb]{0.00,0.07,1.00}{0.8004} & \textcolor[rgb]{1.00,0.00,0.00}{28.96}/0.8001 & \textcolor[rgb]{0.00,0.07,1.00}{28.95}/\textcolor[rgb]{1.00,0.00,0.00}{0.8013} \\
			
			& $\times 4$ & 25.96/0.6675 & 27.29/0.7251 & 27.23/0.7233 & 27.32/0.7275 & 27.38/\textcolor[rgb]{0.00,0.07,1.00}{0.7284} & \textcolor[rgb]{0.00,0.07,1.00}{27.40}/0.7281 & \textcolor[rgb]{1.00,0.00,0.00}{27.41}/\textcolor[rgb]{1.00,0.00,0.00}{0.7297} \\
			
			\hline
			\hline
			
			\multirow{3}{*}{Urban100} & $\times 2$ & 26.88/0.8403 & 30.76/0.9140 & 30.75/0.9133 & 30.41/0.9103 & 31.23/0.9188 & \textcolor[rgb]{1.00,0.00,0.00}{31.31}/\textcolor[rgb]{0.00,0.07,1.00}{0.9195} & \textcolor[rgb]{0.00,0.07,1.00}{31.27}/\textcolor[rgb]{1.00,0.00,0.00}{0.9196} \\
			
			& $\times 3$ & 24.46/0.7349 & 27.14/0.8279 & 27.15/0.8276 & 27.07/0.8275 & \textcolor[rgb]{0.00,0.07,1.00}{27.53}/\textcolor[rgb]{1.00,0.00,0.00}{0.8378} & \textcolor[rgb]{1.00,0.00,0.00}{27.56}/\textcolor[rgb]{0.00,0.07,1.00}{0.8376} & 27.42/0.8359 \\
			
			& $\times 4$ & 23.14/0.6577 & 25.18/0.7524 & 25.14/0.7510 & 25.21/0.7562 & \textcolor[rgb]{0.00,0.07,1.00}{25.44}/\textcolor[rgb]{1.00,0.00,0.00}{0.7638} & \textcolor[rgb]{1.00,0.00,0.00}{25.50}/0.7630 & 25.41/\textcolor[rgb]{0.00,0.07,1.00}{0.7632} \\
			
			\hline
			
		\end{tabular}
	\end{center}
	\caption{Average PSNR/SSIMs for scale $2 \times$, $3 \times$ and $4 \times$. Red color indicates the best and blue color indicates the second best performance.}
	\label{tab:psnr/ssim}
\end{table*}

\begin{table*}[htb]
	\small
	\centering
	\begin{tabular}{|c|c|c|c|c|c|c|c||c|}
		\hline
		Dataset & Scale & Bicubic & VDSR~\cite{VDSR} & DRCN~\cite{DRCN} & LapSRN~\cite{LapSRN} & DRRN~\cite{DRRN} & MemNet~\cite{MemNet} & IDN~(Ours) \\
		\hline
		\hline
		\multirow{3}{*}{Set5} & $\times 2$ & 6.083 & 8.580 & 8.783 & \textcolor[rgb]{0.00,0.07,1.00}{9.010} & 8.670 & 8.850 & \textcolor[rgb]{1.00,0.00,0.00}{9.252} \\
		& $\times 3$ & 3.580 & 5.203 & 5.336 & 5.194 & 5.394 & \textcolor[rgb]{0.00,0.07,1.00}{5.503} & \textcolor[rgb]{1.00,0.00,0.00}{5.620} \\
		& $\times 4$ & 2.329 & 3.542 & 3.543 & 3.559 & 3.700 & \textcolor[rgb]{0.00,0.07,1.00}{3.787} & \textcolor[rgb]{1.00,0.00,0.00}{3.826} \\
		\hline
		\hline
		\multirow{3}{*}{Set14} & $\times 2$ & 6.105 & 8.159 & 8.370 & \textcolor[rgb]{0.00,0.07,1.00}{8.501} & 8.280 & 8.469 & \textcolor[rgb]{1.00,0.00,0.00}{8.839} \\
		& $\times 3$ & 3.473 & 4.691 & 4.782 & 4.662 & 4.870 & \textcolor[rgb]{0.00,0.07,1.00}{4.958} & \textcolor[rgb]{1.00,0.00,0.00}{5.062} \\
		& $\times 4$ & 2.237 & 3.106 & 3.098 & 3.145 & 3.249 & \textcolor[rgb]{0.00,0.07,1.00}{3.309} & \textcolor[rgb]{1.00,0.00,0.00}{3.354} \\
		\hline
		\hline
		
		\multirow{3}{*}{BSD100} & $\times 2$ & 5.619 & 7.494 & 7.577 & \textcolor[rgb]{0.00,0.07,1.00}{7.715} & 7.513 & 7.665 & \textcolor[rgb]{1.00,0.00,0.00}{7.931} \\
		& $\times 3$ & 3.138 & 4.151 & 4.184 & 4.057 & 4.235 & \textcolor[rgb]{0.00,0.07,1.00}{4.300} & \textcolor[rgb]{1.00,0.00,0.00}{4.398} \\
		& $\times 4$ & 1.978 & 2.679 & 2.633 & 2.677 & 2.746 & \textcolor[rgb]{0.00,0.07,1.00}{2.778} & \textcolor[rgb]{1.00,0.00,0.00}{2.837} \\
		\hline
		\hline
		
		\multirow{3}{*}{Urban100} & $\times 2$ & 6.245 & 8.629 & 8.959 & 8.907 & 8.889 & \textcolor[rgb]{0.00,0.07,1.00}{9.122} & \textcolor[rgb]{1.00,0.00,0.00}{9.594} \\
		& $\times 3$ & 3.620 & 5.159 & 5.314 & 5.156 & 5.440 & \textcolor[rgb]{0.00,0.07,1.00}{5.560} & \textcolor[rgb]{1.00,0.00,0.00}{5.676} \\
		& $\times 4$ & 2.361 & 3.462 & 3.465 & 3.530 & 3.669 & \textcolor[rgb]{0.00,0.07,1.00}{3.786} & \textcolor[rgb]{1.00,0.00,0.00}{3.789} \\
		\hline
	\end{tabular}
	\caption{Average IFCs for scale $2 \times$, $3 \times$ and $4 \times$. Red color indicates the best and blue color indicates the second best performance.}
	\label{tab:ifc}
\end{table*}

\begin{table*}[htb]
	\small
	\centering
	\begin{tabular}{|c|c|c|c|c|c|c||c|}
		\hline
		Dataset & Scale & VDSR~\cite{VDSR} & DRCN~\cite{DRCN} & LapSRN~\cite{LapSRN} & DRRN~\cite{DRRN} & MemNet~\cite{MemNet} & IDN~(Ours) \\
		\hline
		\hline
		\multirow{3}{*}{Set5} & $\times 2$ & 0.054 & 0.735 & \textcolor[rgb]{0.00,0.07,1.00}{0.032} & 4.343 & 5.715 & \textcolor[rgb]{1.00,0.00,0.00}{0.016} \\
		& $\times 3$ & 0.062 & 0.748 & \textcolor[rgb]{0.00,0.07,1.00}{0.049} & 4.380 & 5.761 & \textcolor[rgb]{1.00,0.00,0.00}{0.011} \\
		& $\times 4$ & 0.054 & 0.735 & \textcolor[rgb]{0.00,0.07,1.00}{0.040} & 4.450 & 5.728 & \textcolor[rgb]{1.00,0.00,0.00}{0.009} \\
		\hline
		\hline
		\multirow{3}{*}{Set14} & $\times 2$ & 0.113 & 1.579 & \textcolor[rgb]{0.00,0.07,1.00}{0.035} & 8.540 & 12.031 & \textcolor[rgb]{1.00,0.00,0.00}{0.025} \\
		& $\times 3$ & 0.122 & 1.569 & \textcolor[rgb]{0.00,0.07,1.00}{0.061} & 8.298 & 11.543 & \textcolor[rgb]{1.00,0.00,0.00}{0.014} \\
		
		& $\times 4$ & 0.112 & 1.526 & \textcolor[rgb]{0.00,0.07,1.00}{0.040} & 8.540 & 11.956 & \textcolor[rgb]{1.00,0.00,0.00}{0.010} \\
		\hline
		\hline
		\multirow{3}{*}{BSD100} & $\times 2$ & 0.071 & 0.983 & \textcolor[rgb]{0.00,0.07,1.00}{0.018} & 4.430 & 5.875 & \textcolor[rgb]{1.00,0.00,0.00}{0.015} \\
		& $\times 3$ & 0.071 & 0.996 & \textcolor[rgb]{0.00,0.07,1.00}{0.037} & 4.430 & 5.897 & \textcolor[rgb]{1.00,0.00,0.00}{0.009} \\
		& $\times 4$ & 0.071 & 0.984 & \textcolor[rgb]{0.00,0.07,1.00}{0.023} & 4.373 & 5.887 & \textcolor[rgb]{1.00,0.00,0.00}{0.007} \\
		\hline
		\hline
		\multirow{3}{*}{Urban100} & $\times 2$ & 0.451 & 5.010 & \textcolor[rgb]{0.00,0.07,1.00}{0.082} & 26.699 & 35.871 & \textcolor[rgb]{1.00,0.00,0.00}{0.062} \\
		& $\times 3$ & 0.514 & 5.054 & \textcolor[rgb]{0.00,0.07,1.00}{0.122} & 26.693 & 35.803 & \textcolor[rgb]{1.00,0.00,0.00}{0.034} \\
		
		& $\times 4$ & 0.448 & 5.048 & \textcolor[rgb]{0.00,0.07,1.00}{0.100} & 26.702 & 37.404 & \textcolor[rgb]{1.00,0.00,0.00}{0.022} \\
		
		\hline
	\end{tabular}
	\caption{Comparison the running time (sec) on the 4 benchmark datasets with scale factors $2 \times$, $3 \times$ and $4 \times$. Red color indicates the fastest algorithm and blue color indicates the second fastest method. Our IDN achieves the best time performance.}
	\label{tab:time}
\end{table*}

\begin{figure*}[htb]
	\centering
	\subfigure[Original (PSNR/SSIM/IFC)]{
		\includegraphics[width=0.22\textwidth]{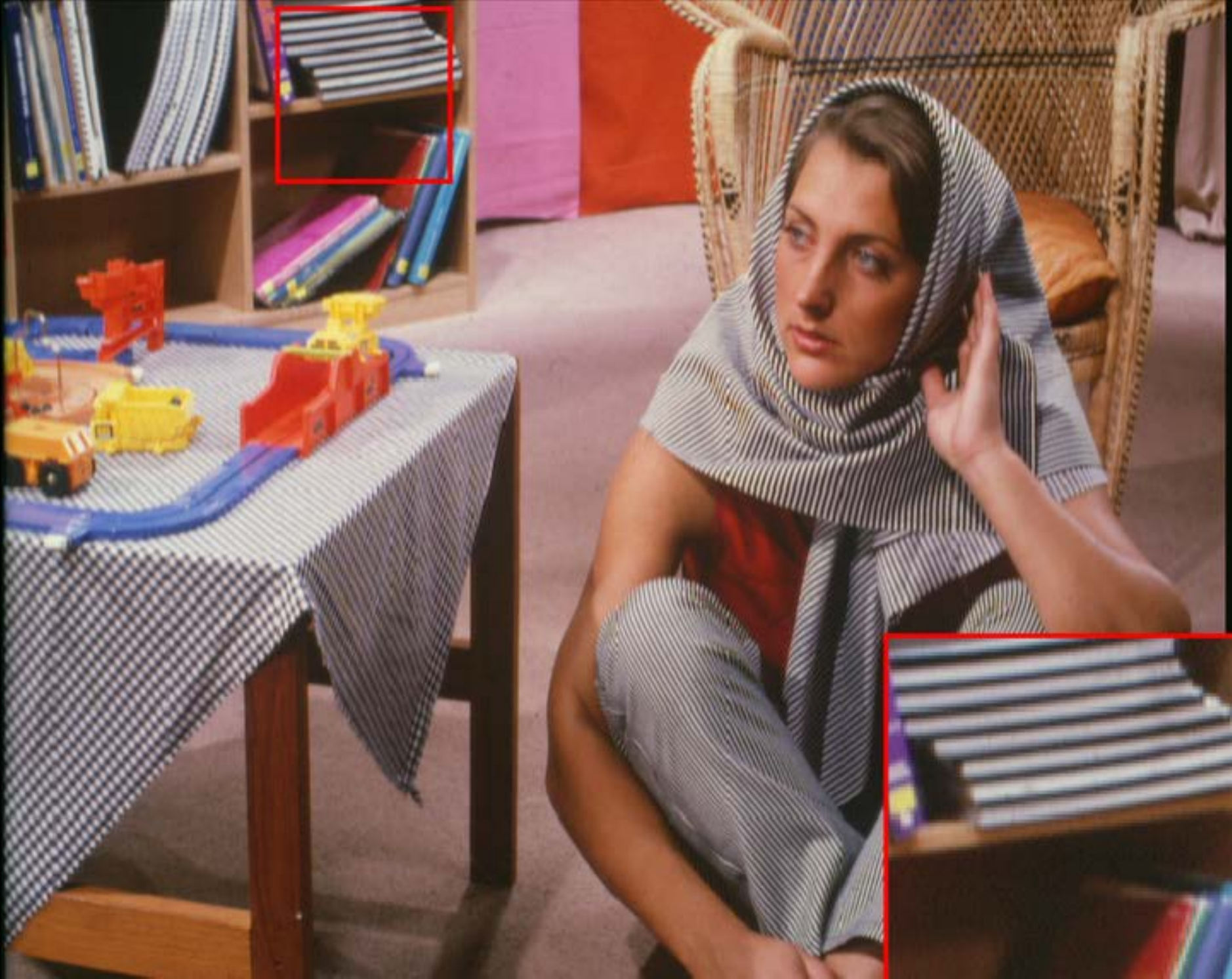}}
	\subfigure[Bicubic (25.15/0.6863/2.553)]{
		\includegraphics[width=0.22\textwidth]{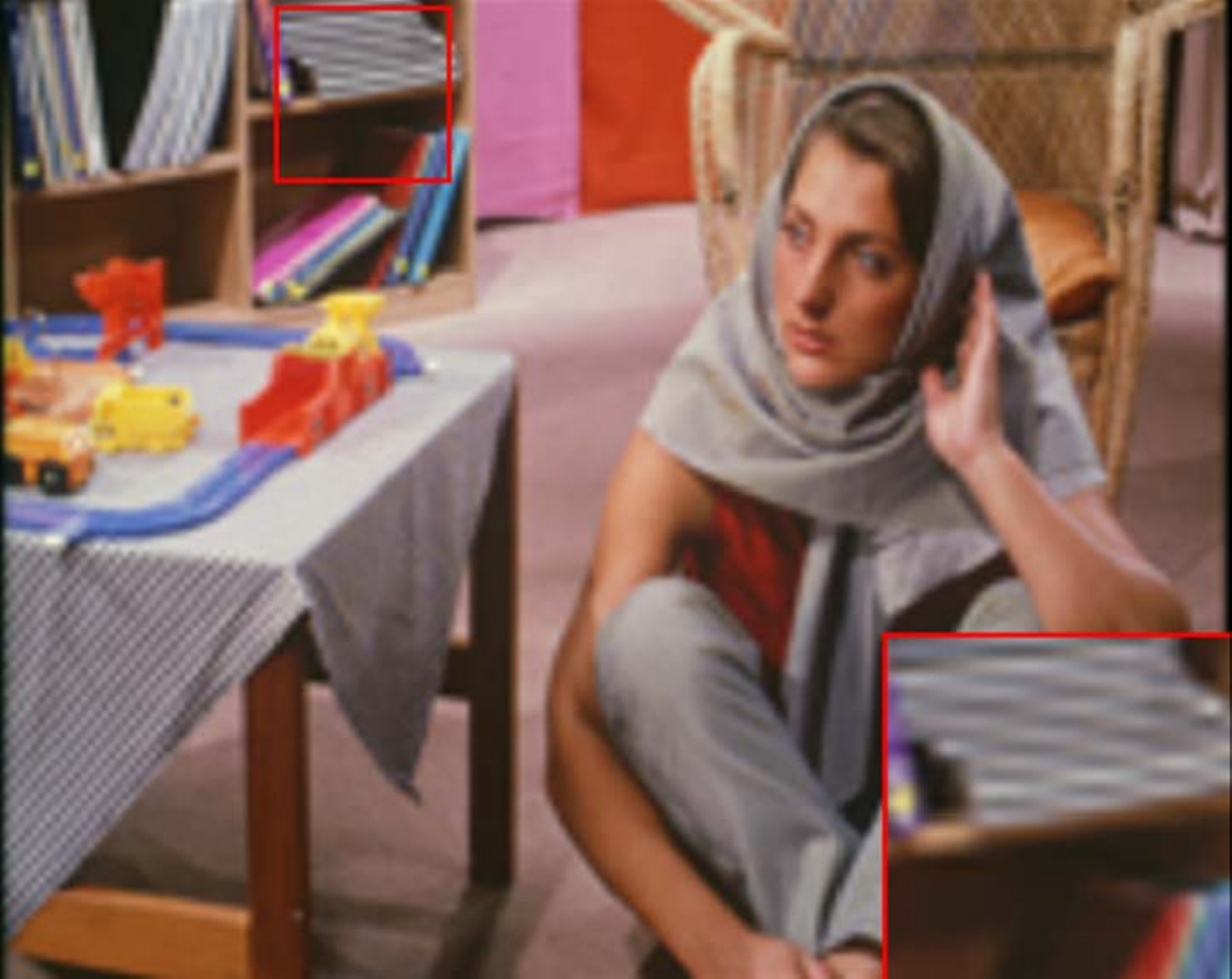}}
	\subfigure[VDSR (25.79/0.7403/3.473)]{
		\includegraphics[width=0.22\textwidth]{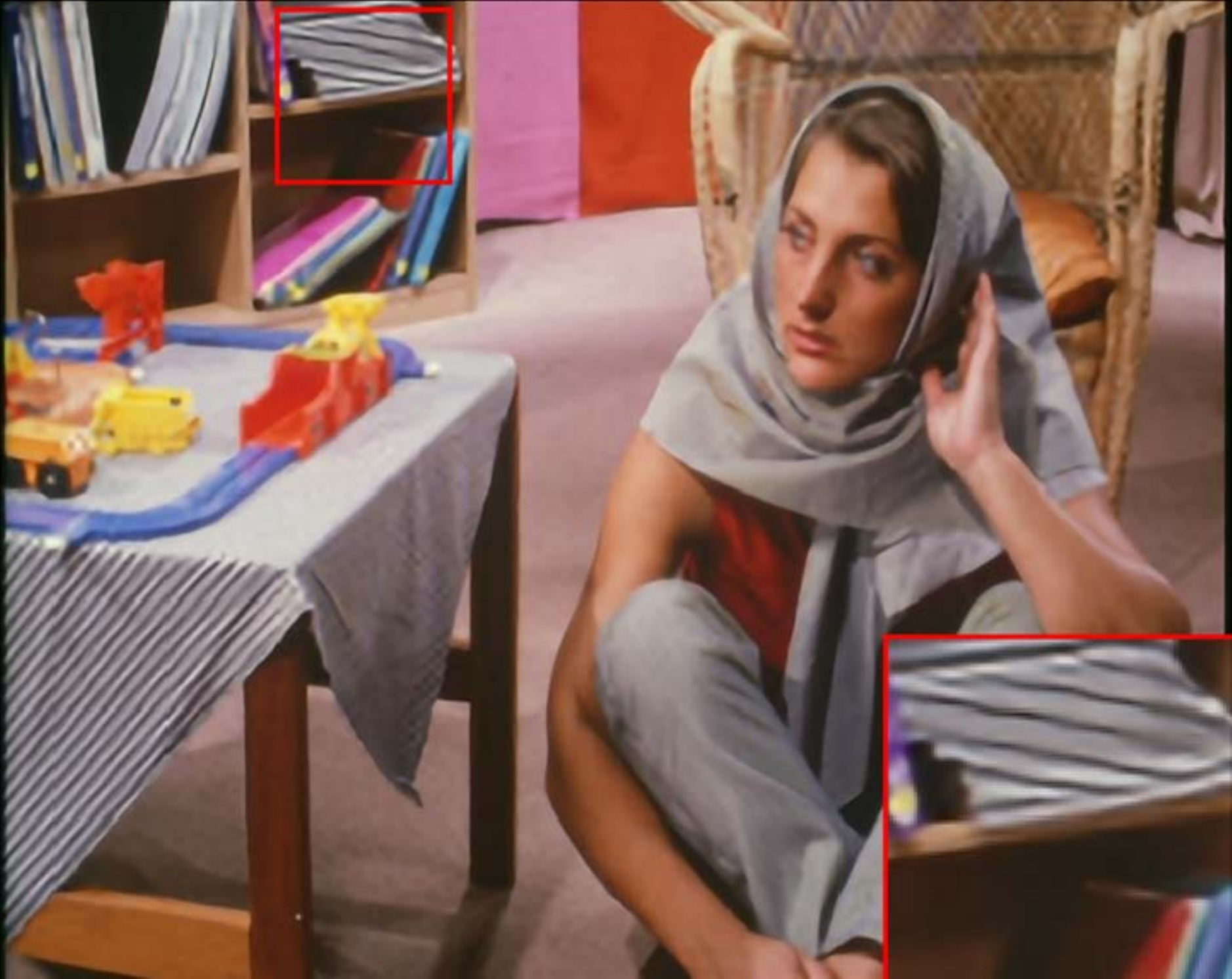}}
	\subfigure[DRCN (25.82/0.7339/3.466)]{
		\includegraphics[width=0.22\textwidth]{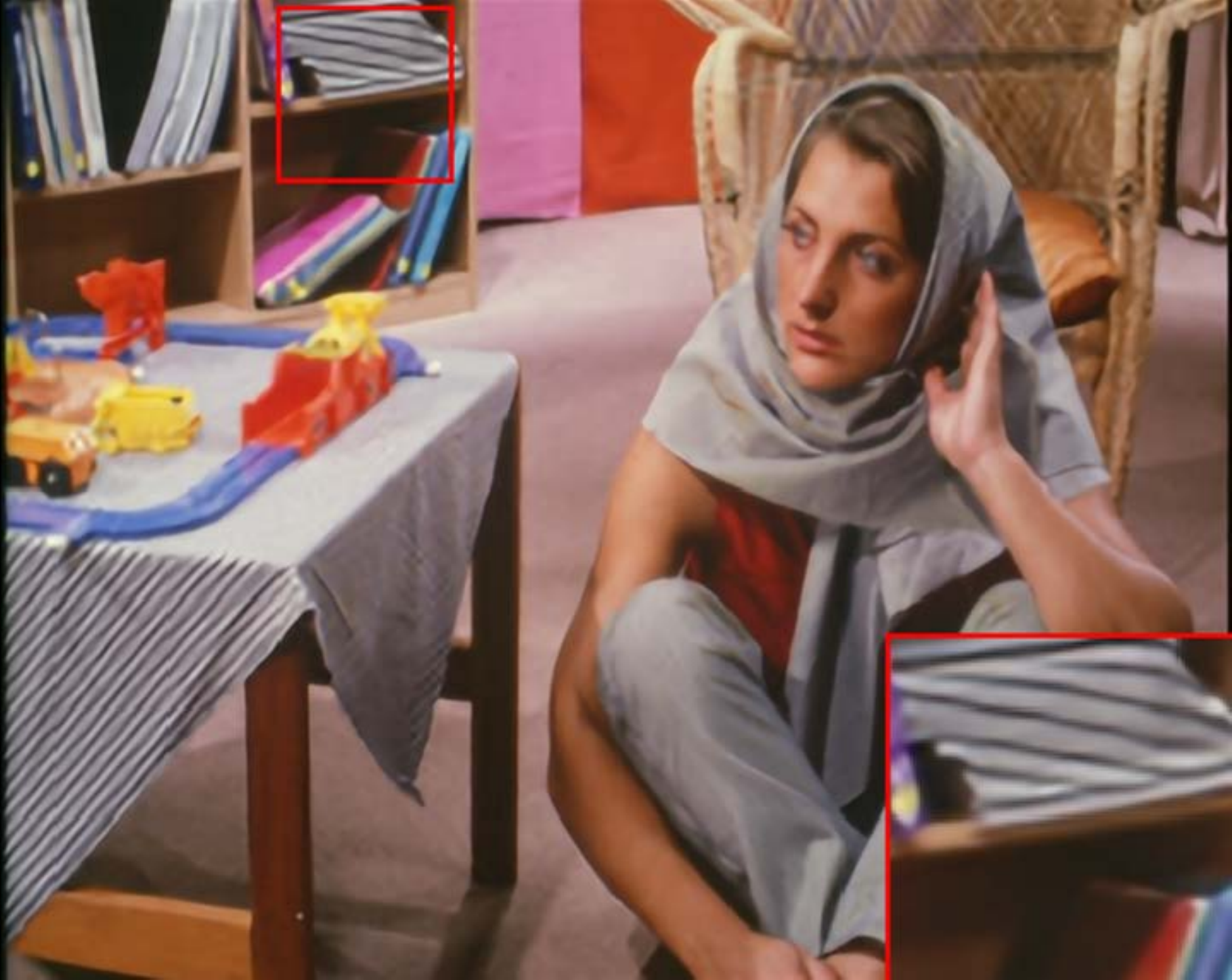}}
	\subfigure[LapSRN (25.77/0.7430/3.546)]{
		\includegraphics[width=0.22\textwidth]{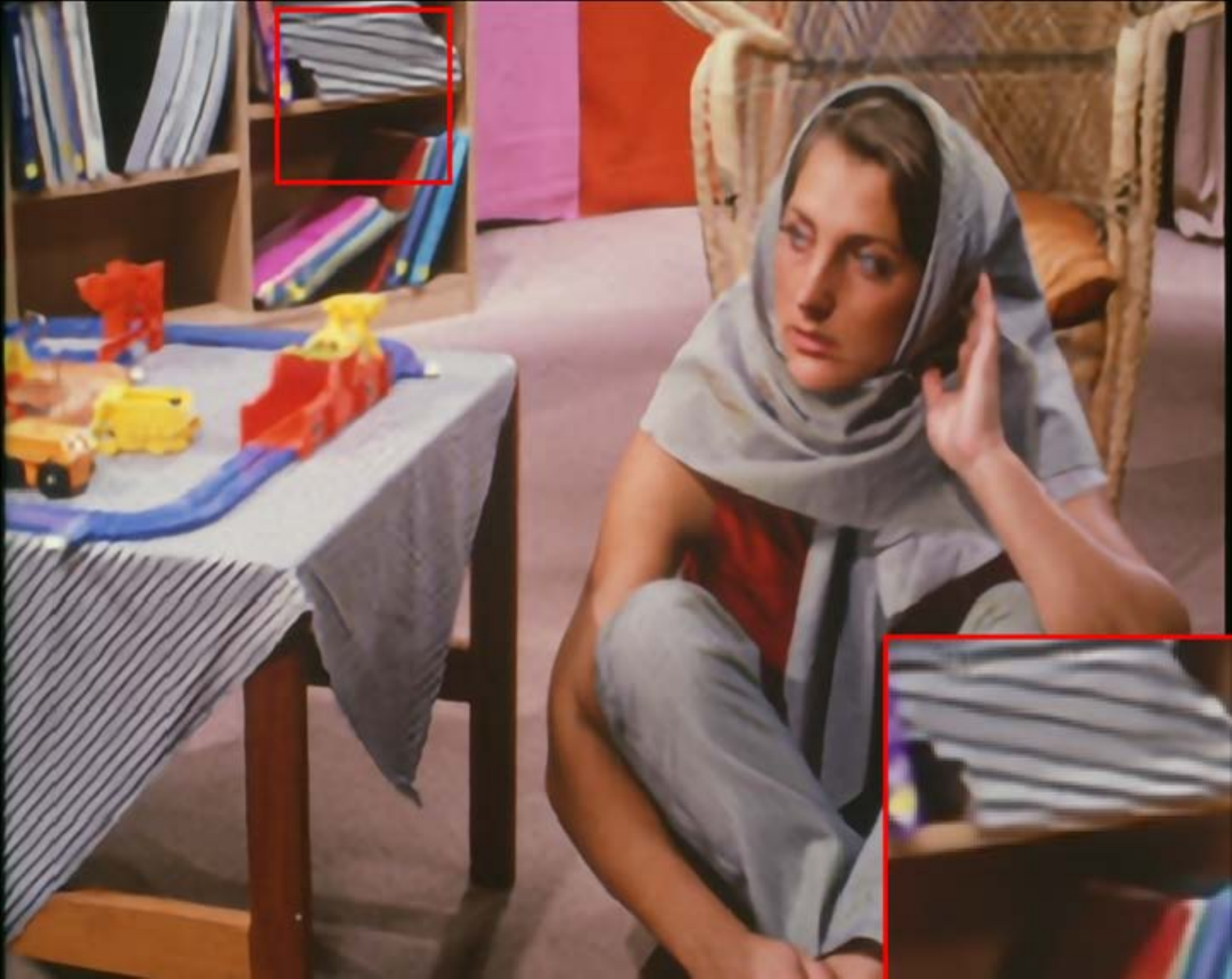}}
	\subfigure[DRRN (25.74/0.7403/3.589)]{
		\includegraphics[width=0.22\textwidth]{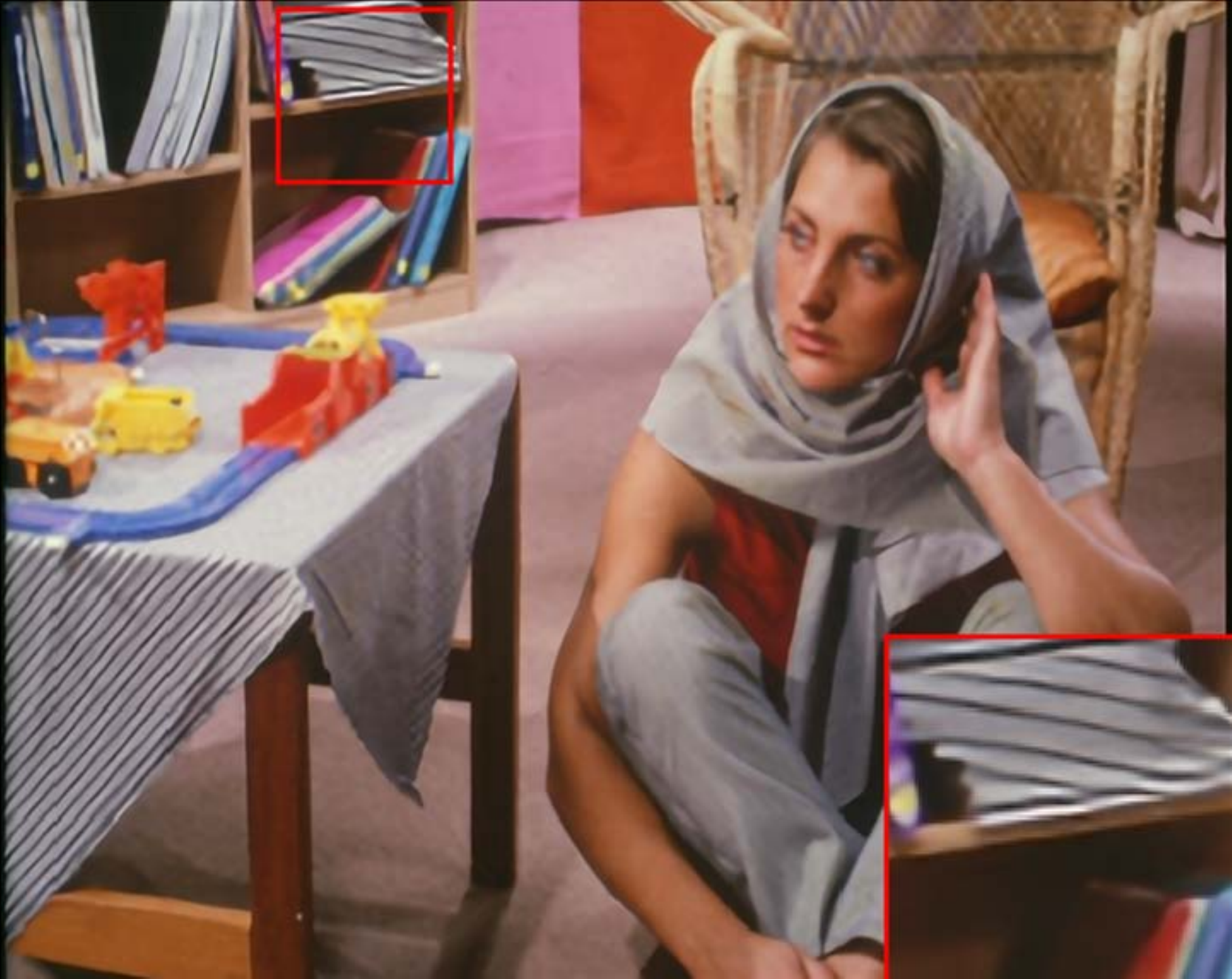}}
	\subfigure[MemNet (25.61/0.7399/3.639)]{
		\includegraphics[width=0.22\textwidth]{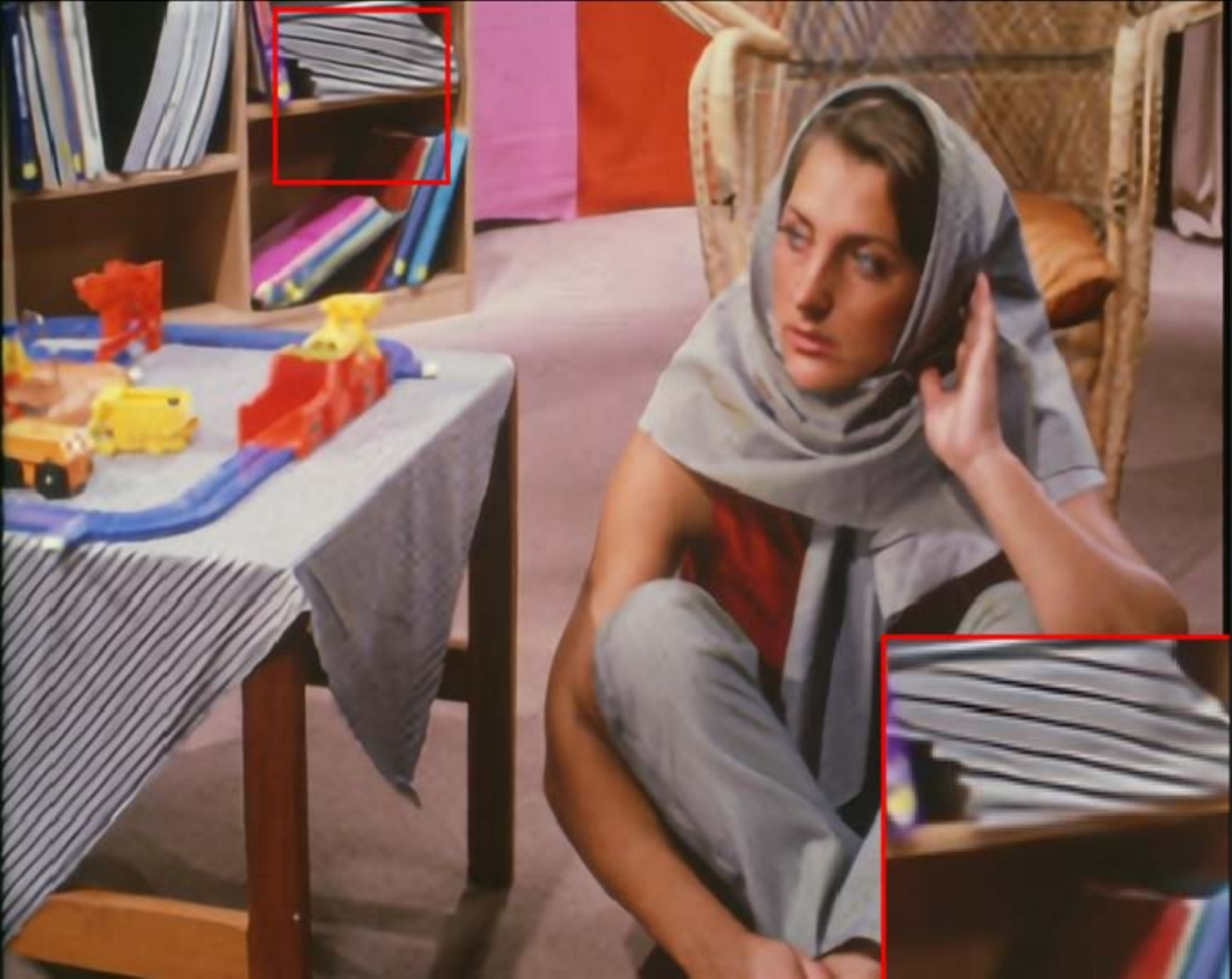}}
	\subfigure[IDN (\textbf{25.84}/\textbf{0.7442}/\textbf{3.749})]{
		\includegraphics[width=0.22\textwidth]{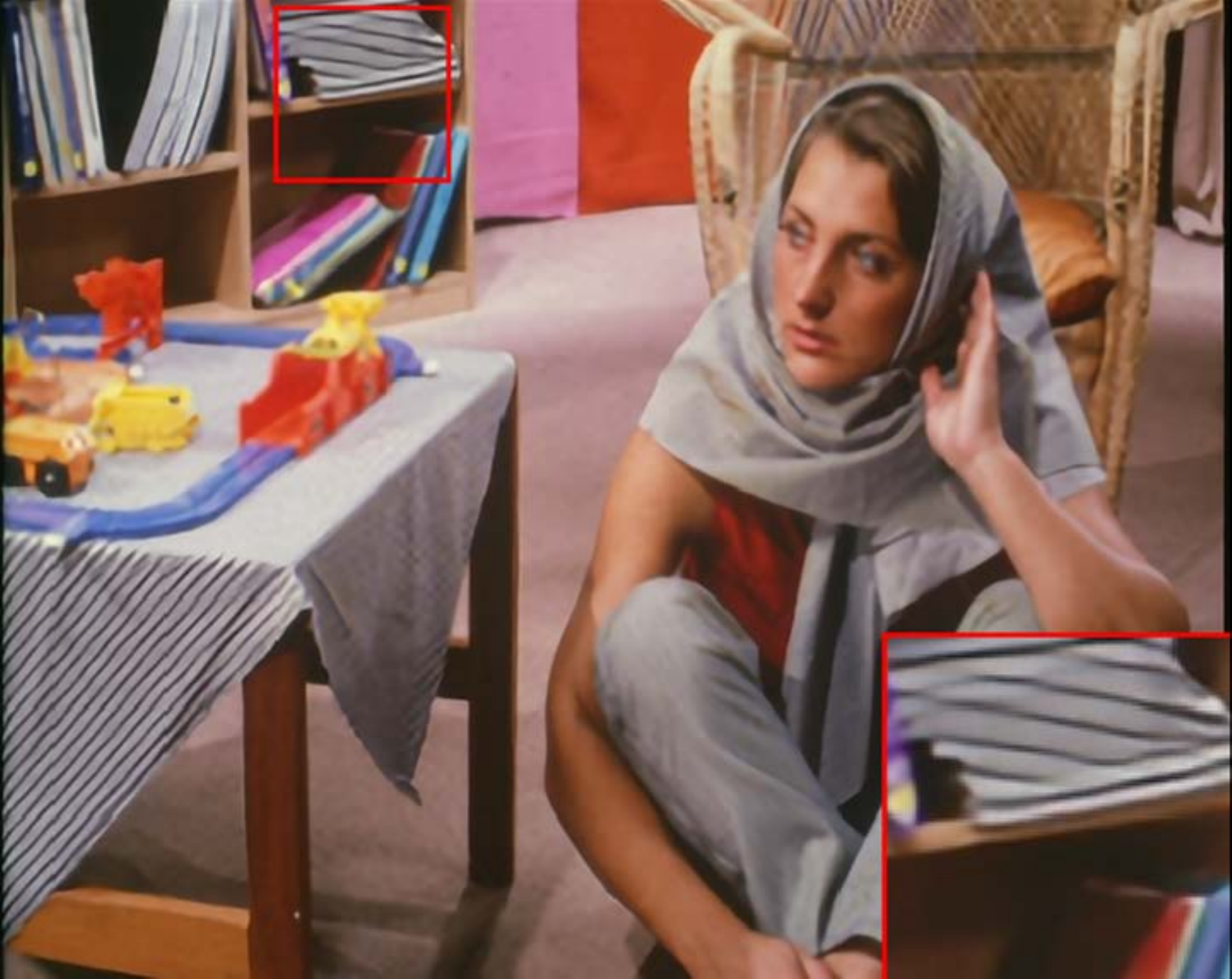}}
	\caption{The ``barbara" image from the Set14 dataset with an upscaling factor 4.}
	\label{fig:barbara}
\end{figure*}

\begin{figure*}[htb]
	\centering
	\subfigure[Original (PSNR/SSIM/IFC)]{
		\includegraphics[width=0.22\textwidth]{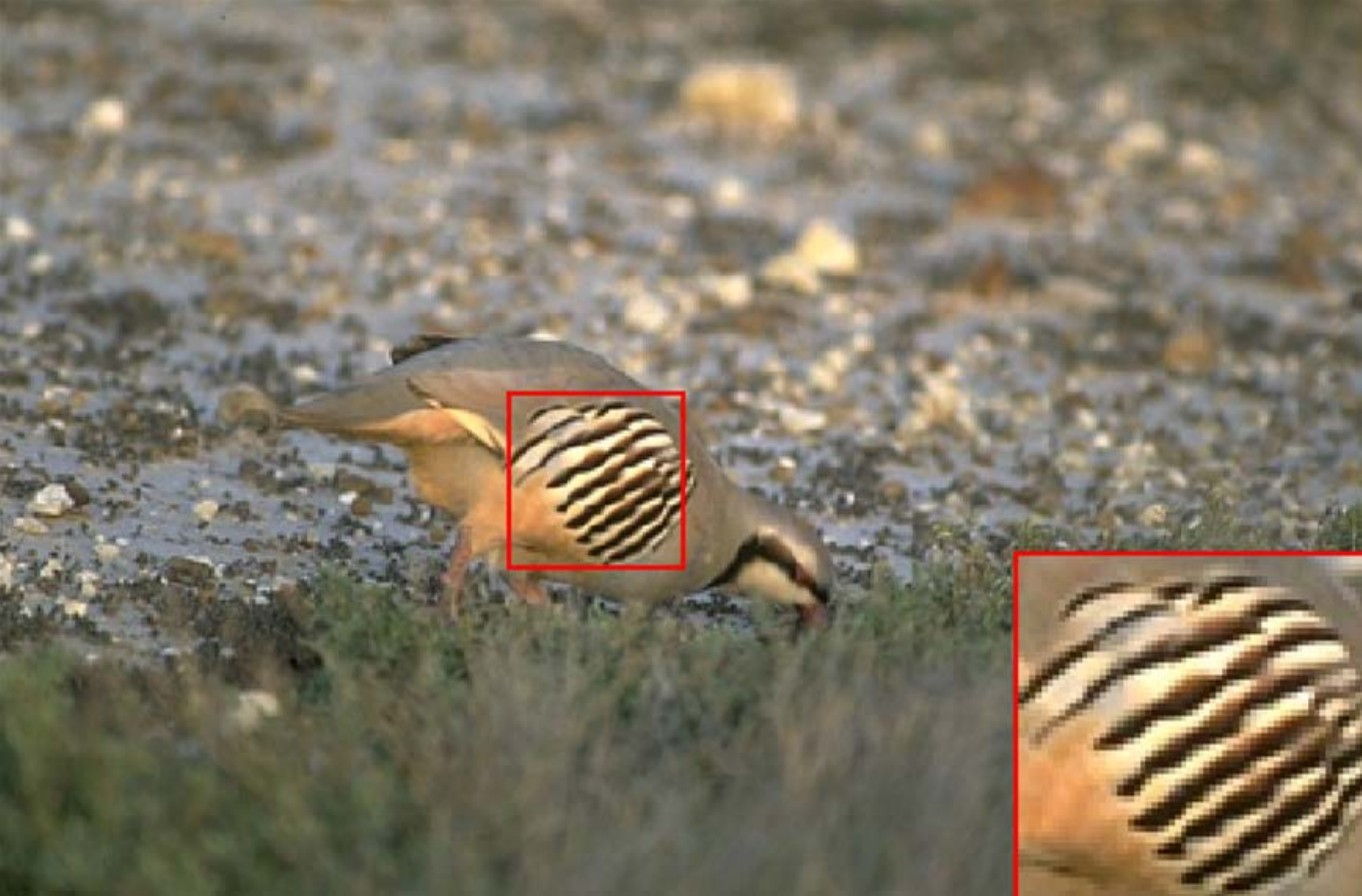}}
	\subfigure[Bicubic (28.50/0.8285/2.638)]{
		\includegraphics[width=0.22\textwidth]{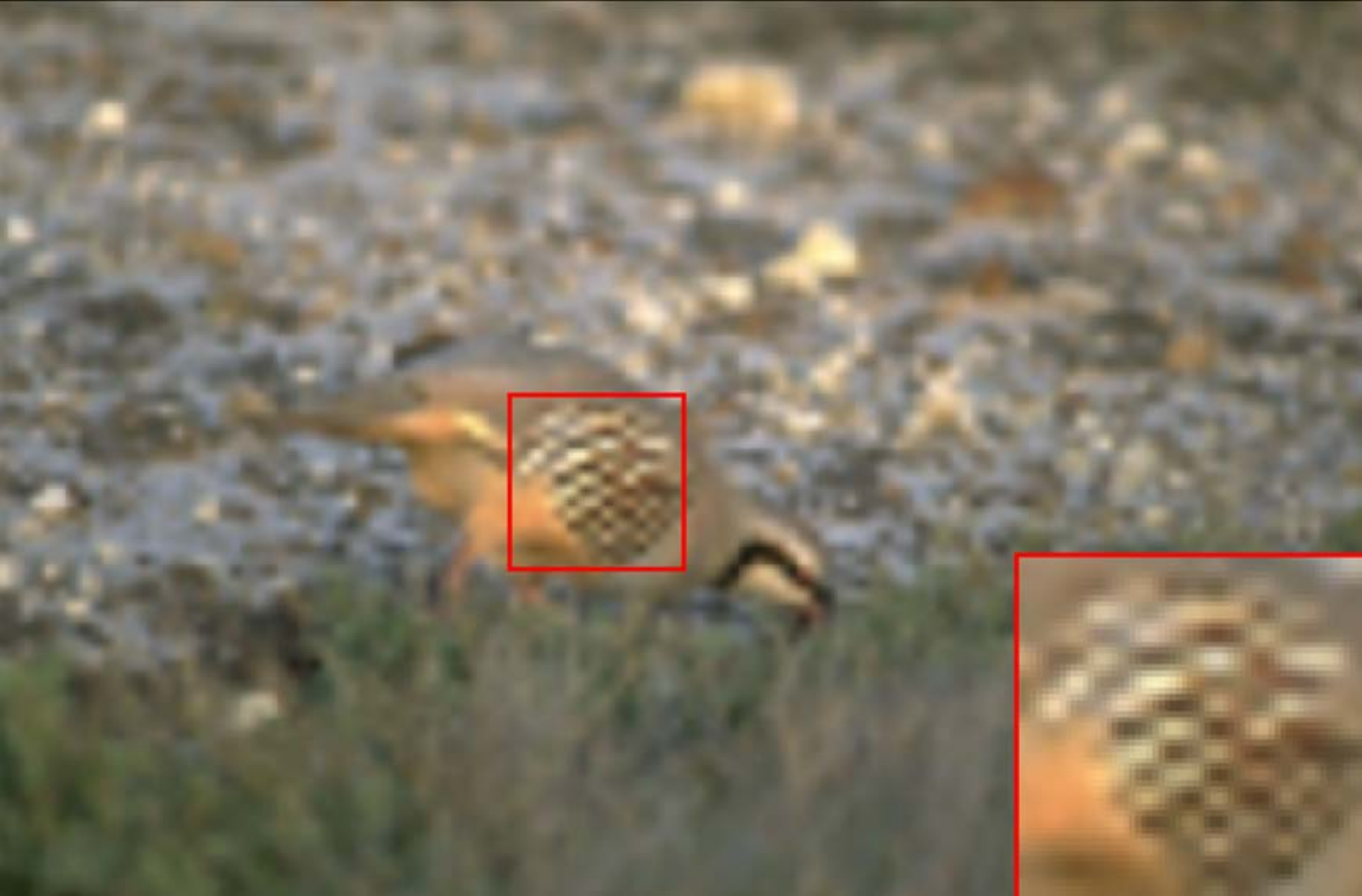}}
	\subfigure[VDSR (29.54/0.8651/3.388)]{
		\includegraphics[width=0.22\textwidth]{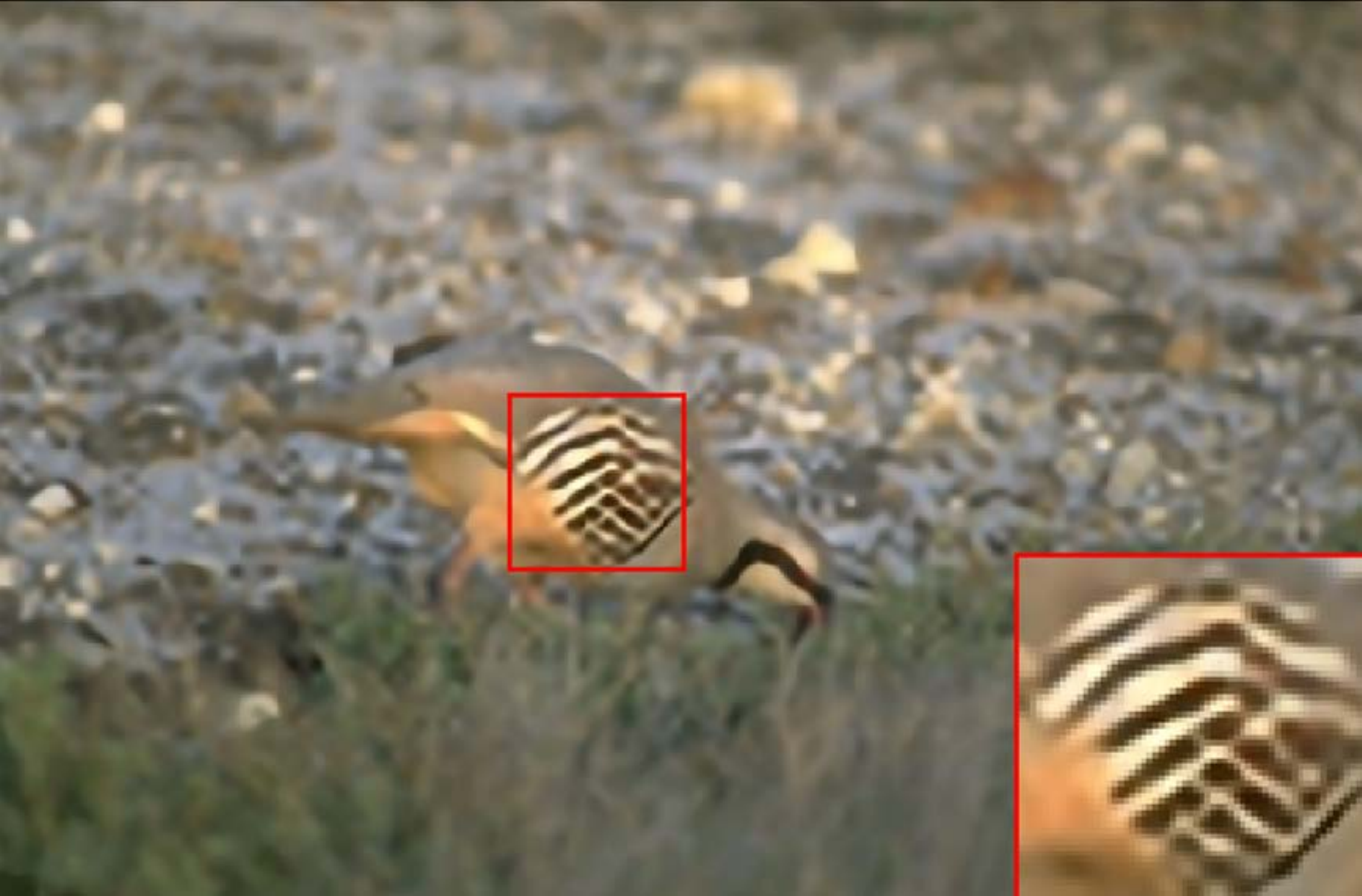}}
	\subfigure[DRCN (30.28/0.8653/3.323)]{
		\includegraphics[width=0.22\textwidth]{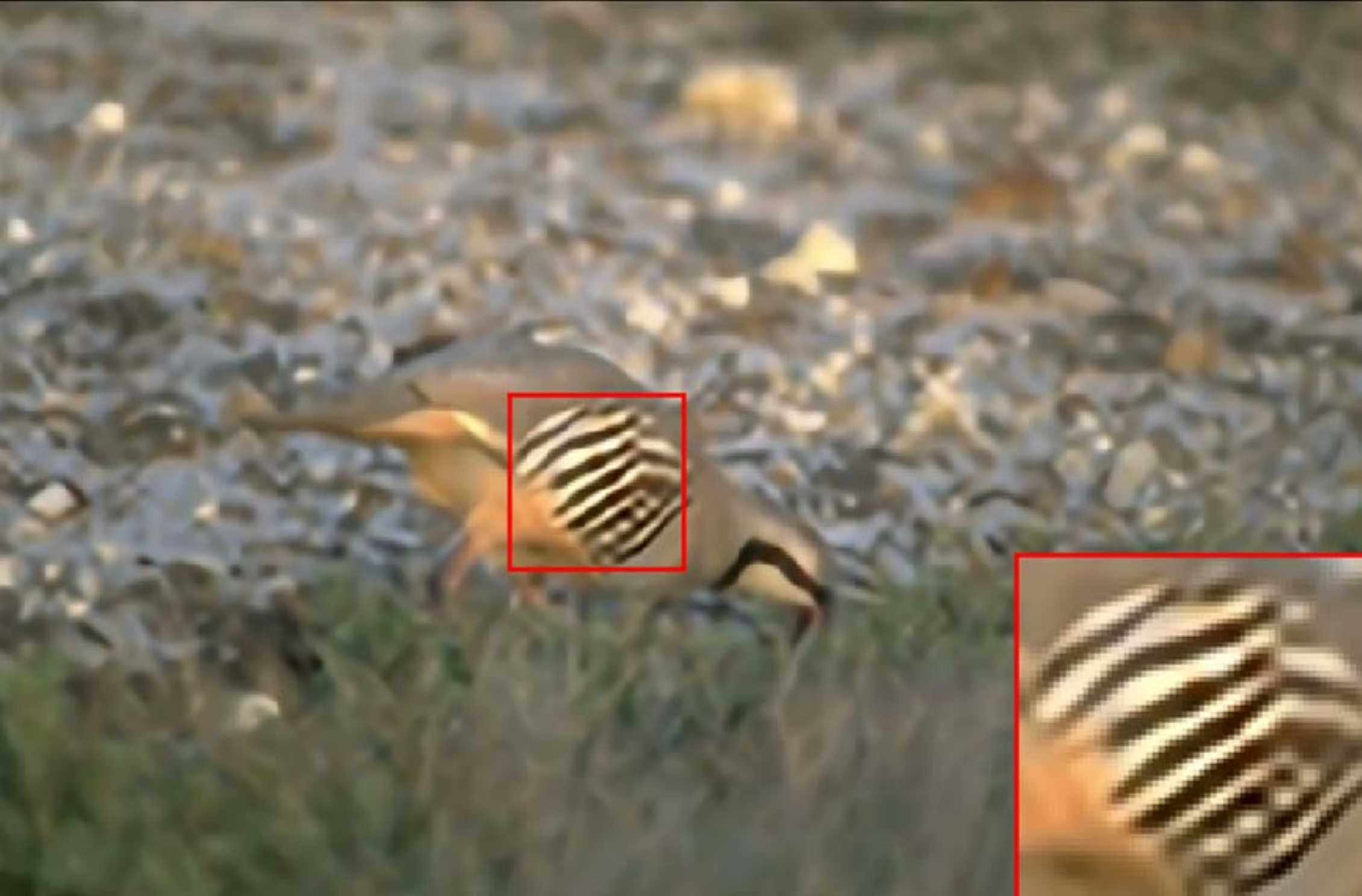}}
	\subfigure[LapSRN (30.10/0.8710/3.432)]{
		\includegraphics[width=0.22\textwidth]{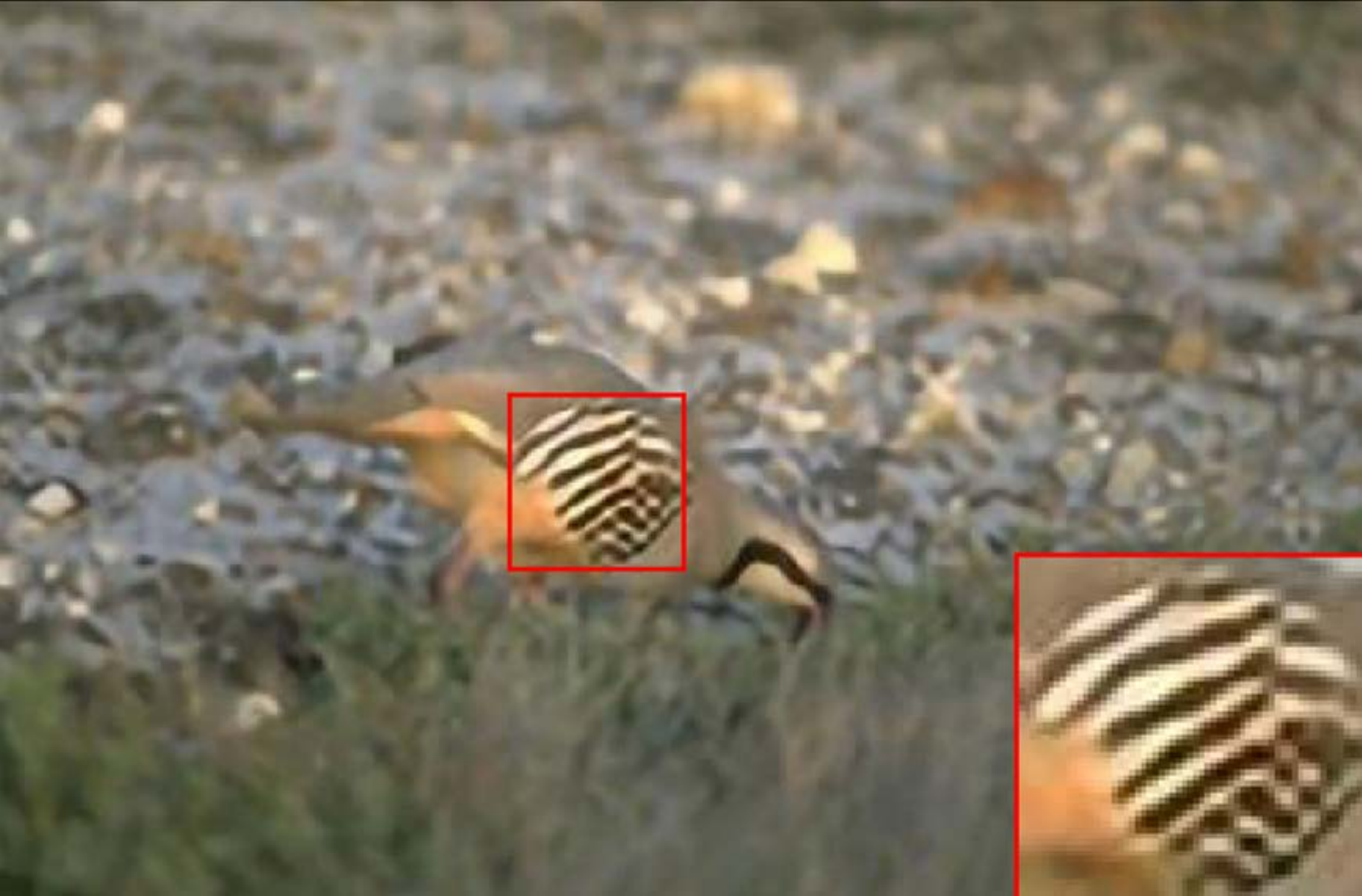}}
	\subfigure[DRRN (29.74/0.8671/3.509)]{
		\includegraphics[width=0.22\textwidth]{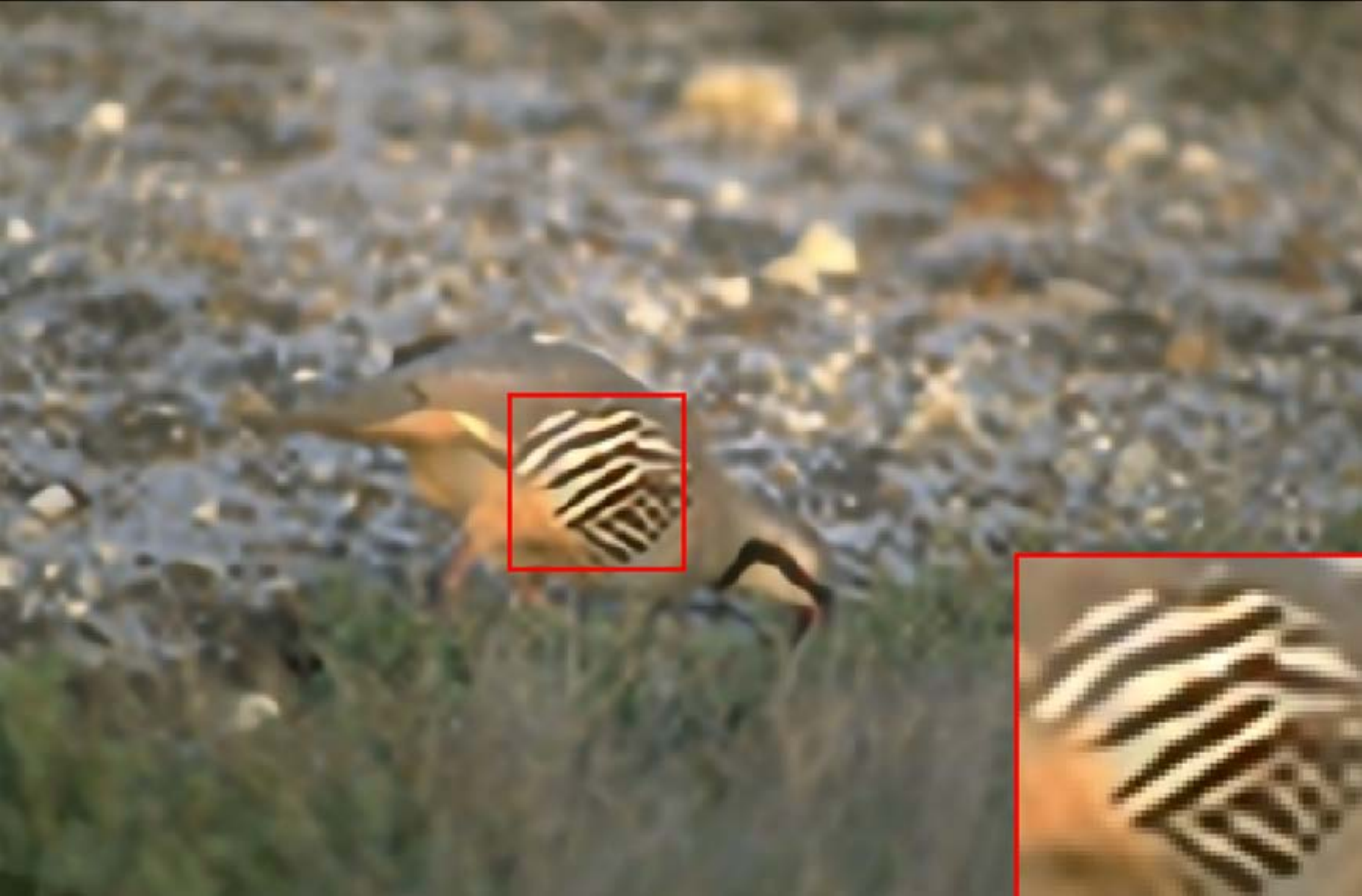}}
	\subfigure[MemNet (30.14/0.8697/3.563)]{
		\includegraphics[width=0.22\textwidth]{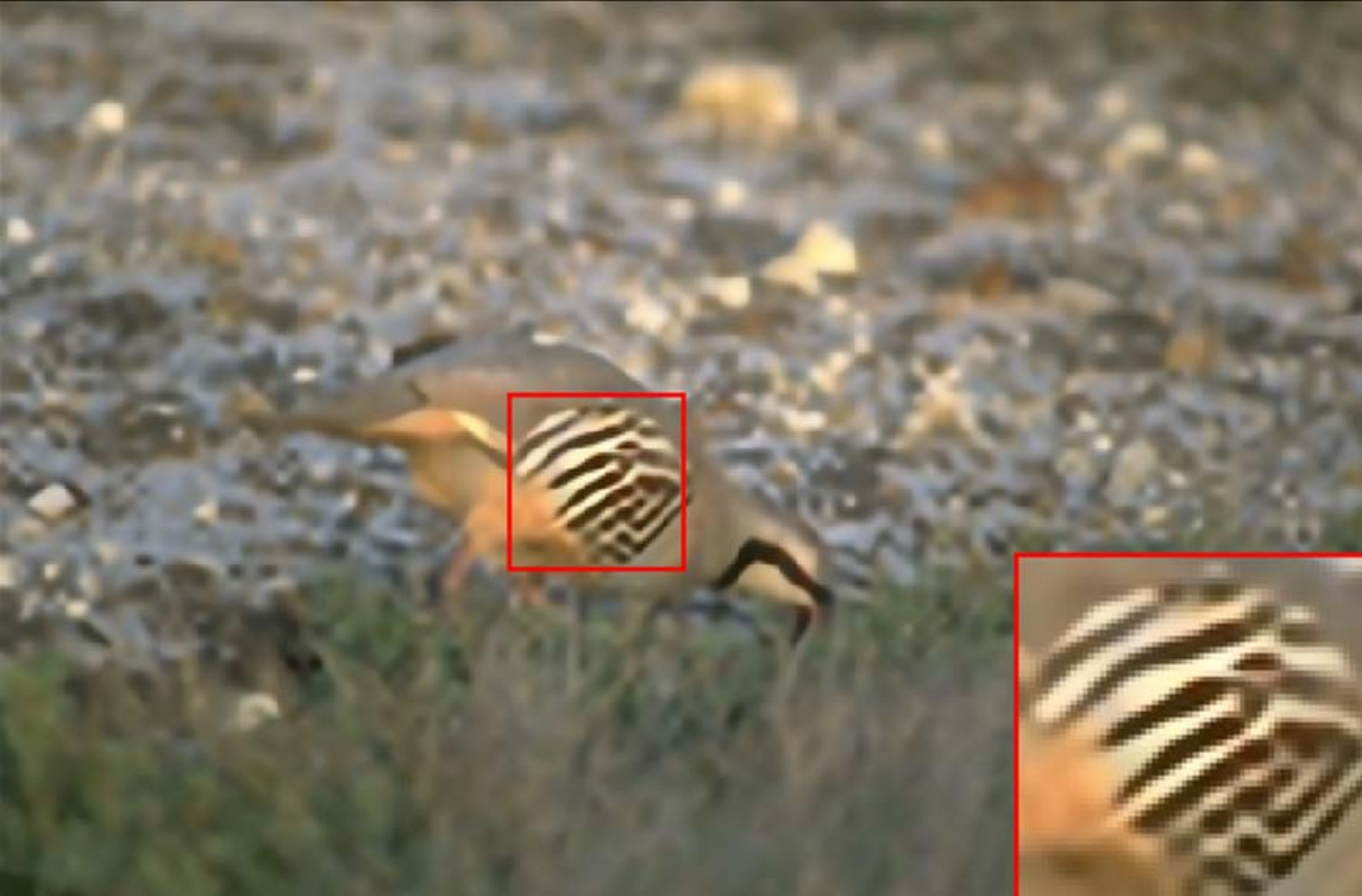}}
	\subfigure[IDN (\textbf{30.40}/\textbf{0.8715}/\textbf{3.703})]{
		\includegraphics[width=0.22\textwidth]{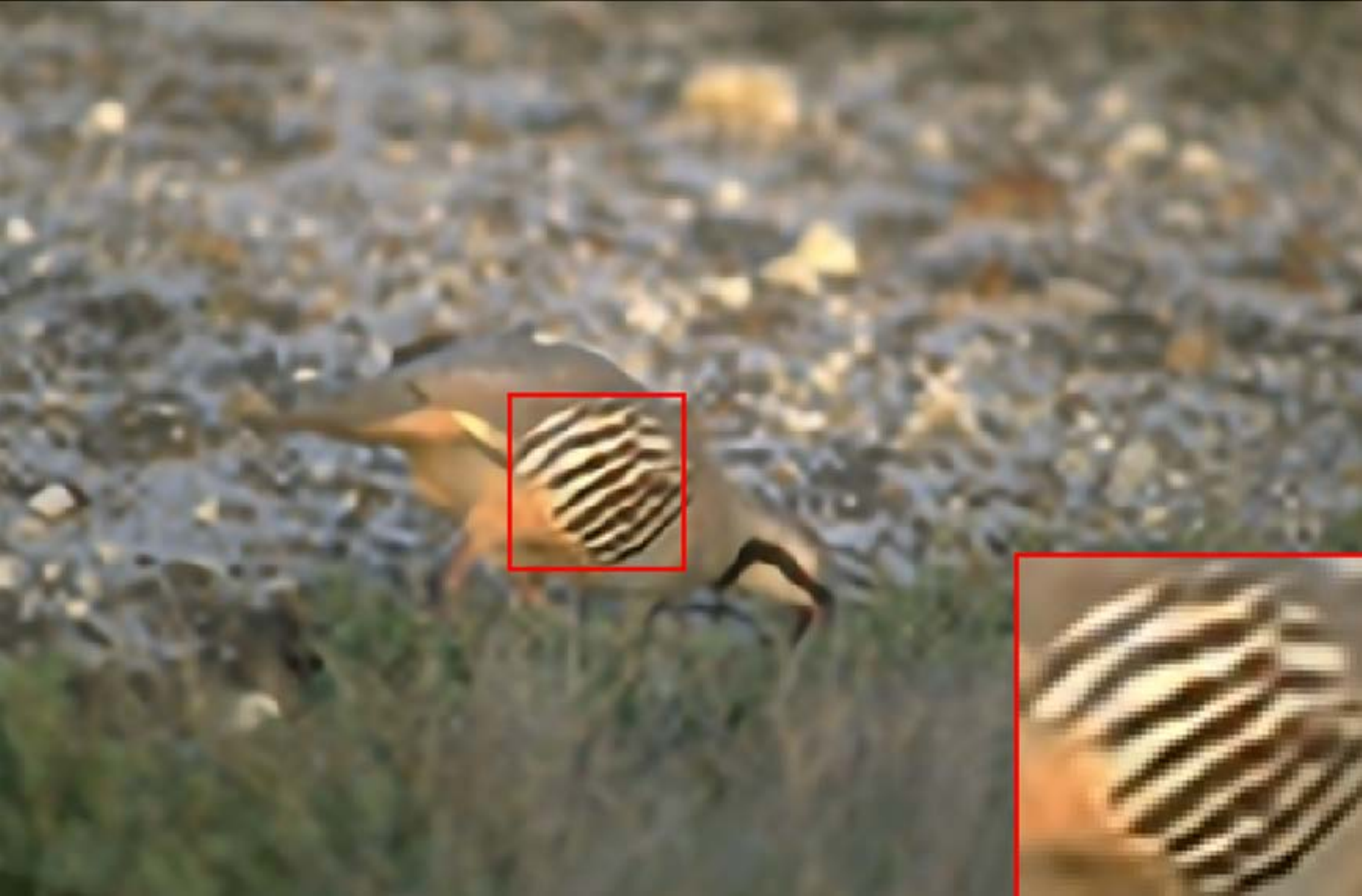}}
	\caption{The ``8023" image from the BSD100 dataset with an upscaling factor 4.}
	\label{fig:8023}
\end{figure*}

\begin{figure*}[htb]
	\centering
	\subfigure[Original (PSNR/SSIM/IFC)]{
		\includegraphics[width=0.22\textwidth]{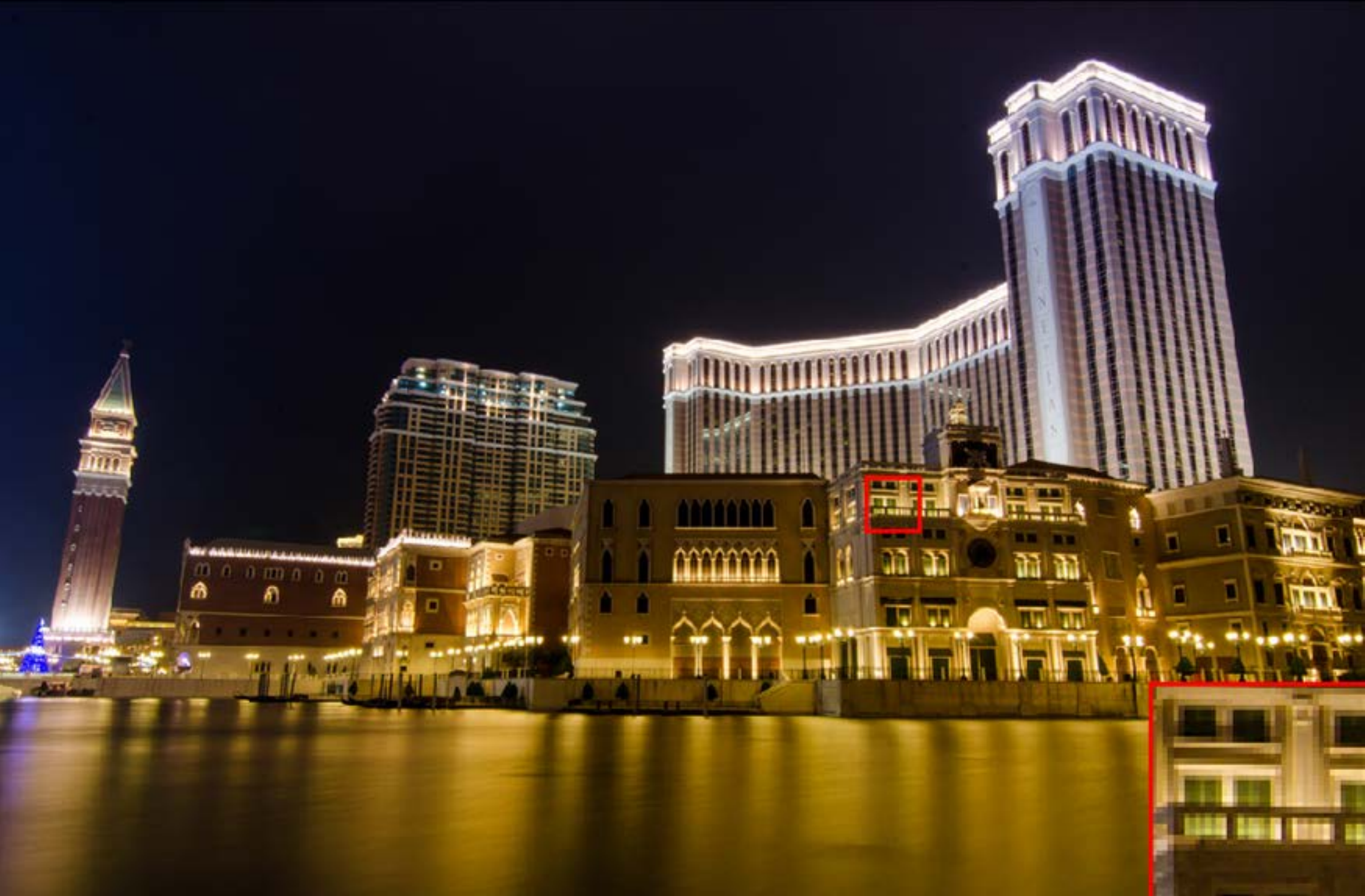}}
	\subfigure[Bicubic (25.90/0.8365/1.976)]{
		\includegraphics[width=0.22\textwidth]{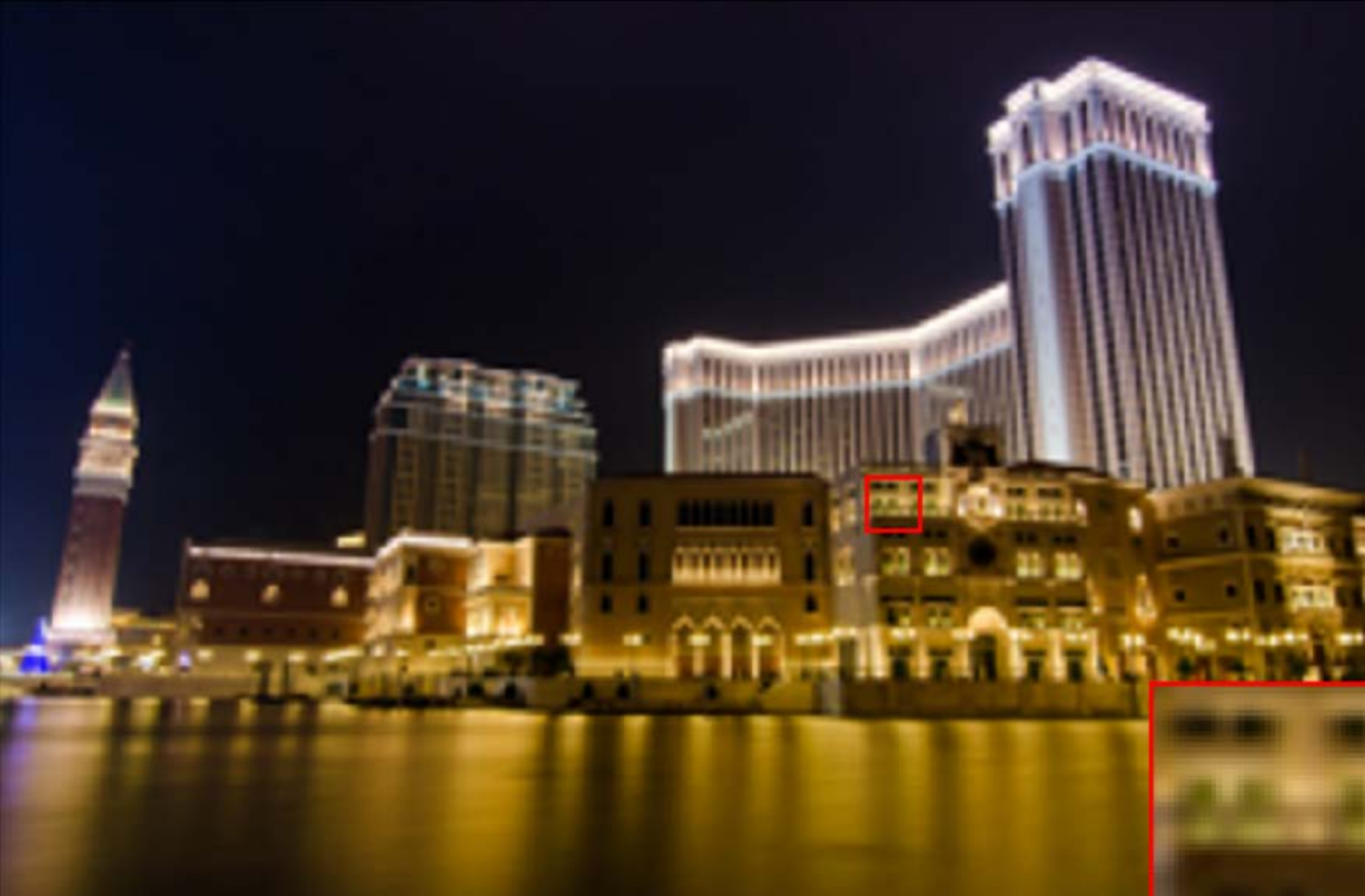}}
	\subfigure[VDSR (27.14/0.8771/2.480)]{
		\includegraphics[width=0.22\textwidth]{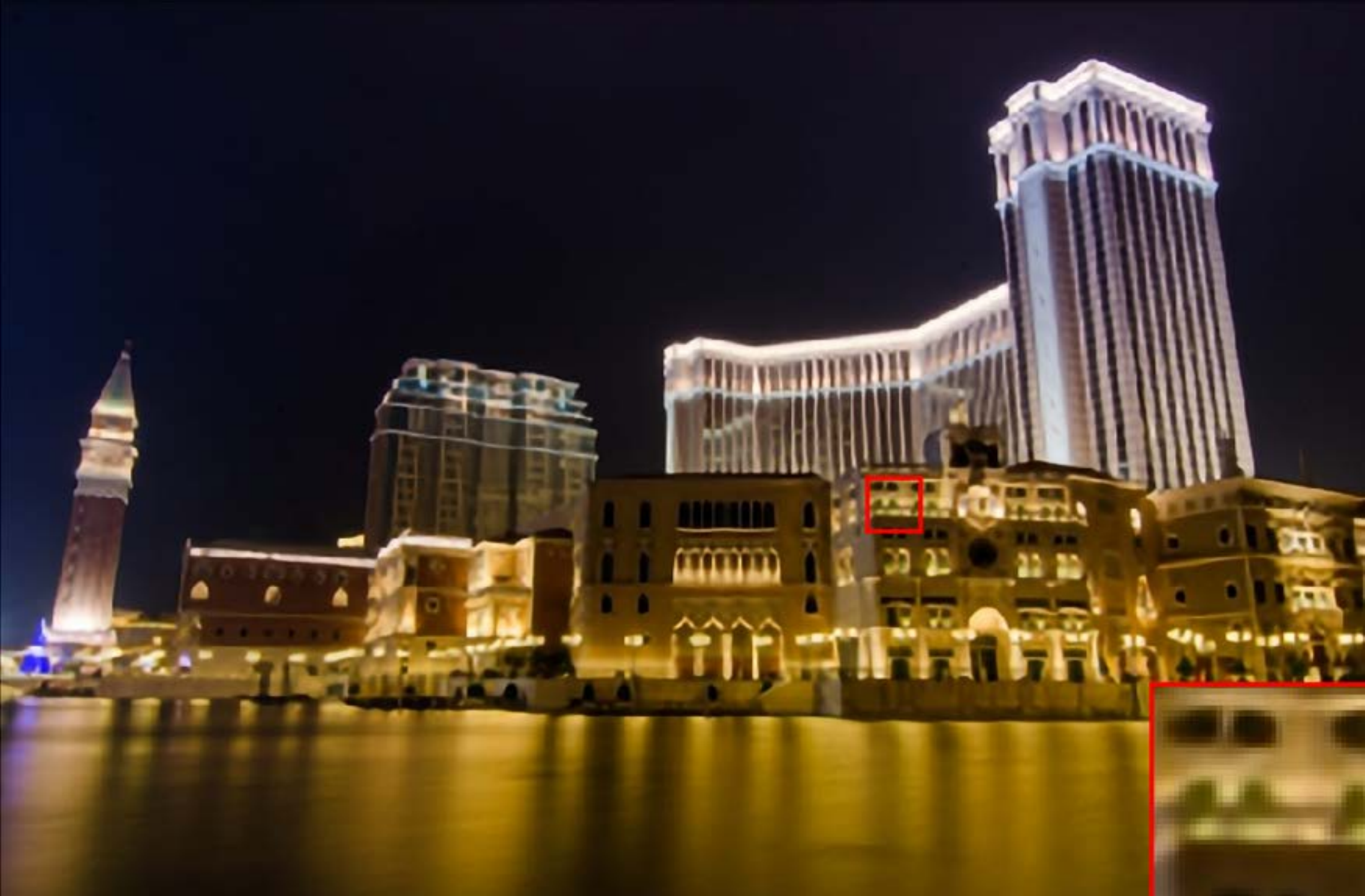}}
	\subfigure[DRCN (27.15/0.8761/2.442)]{
		\includegraphics[width=0.22\textwidth]{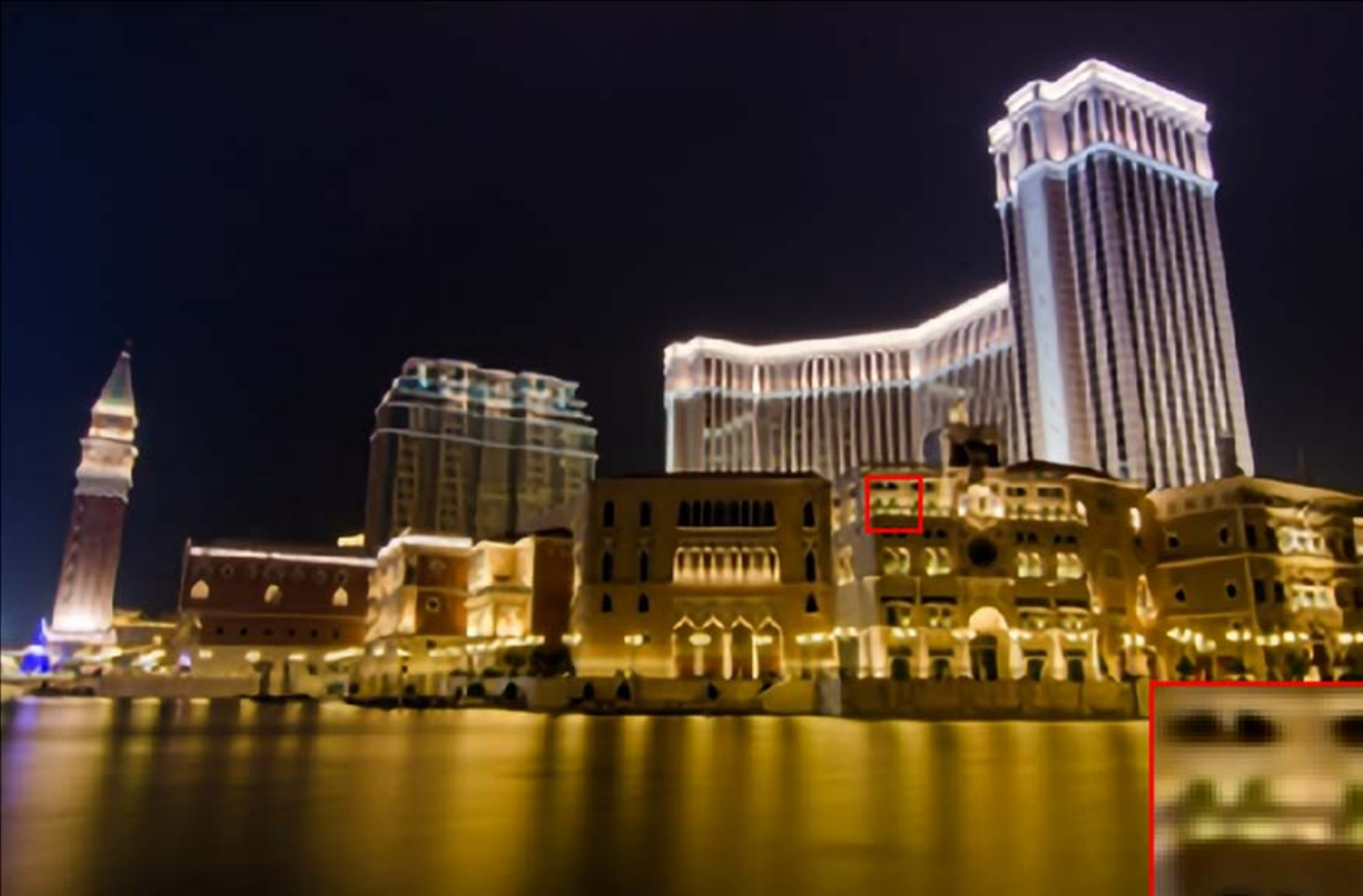}}
	\subfigure[LapSRN (27.11/0.8809/2.593)]{
		\includegraphics[width=0.22\textwidth]{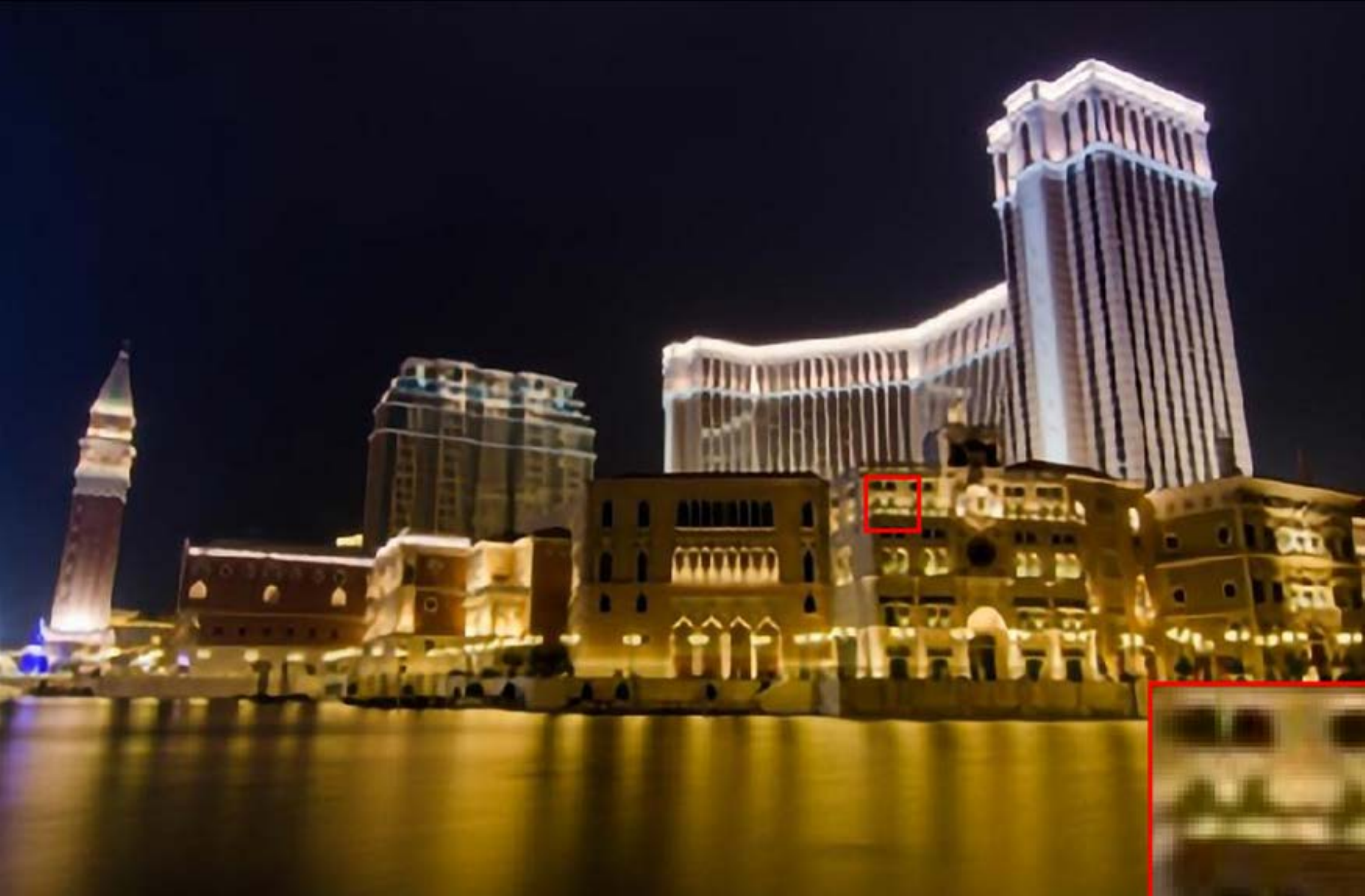}}
	\subfigure[DRRN (26.65/0.8739/2.399)]{
		\includegraphics[width=0.22\textwidth]{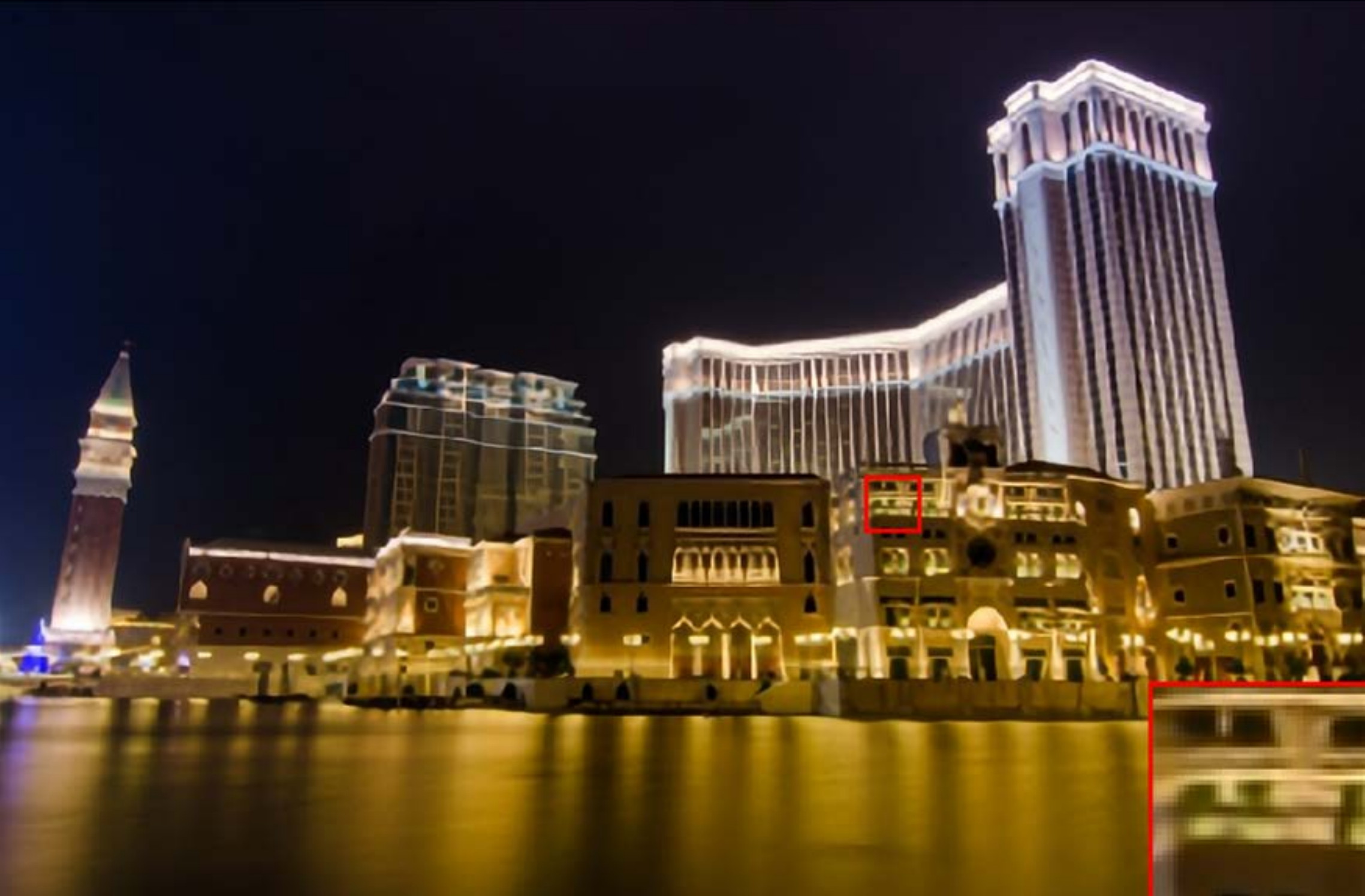}}
	\subfigure[MemNet (26.83/0.8750/2.444)]{
		\includegraphics[width=0.22\textwidth]{figure/img085_drrn.pdf}}
	\subfigure[IDN (\textbf{27.26}/\textbf{0.8824}/\textbf{2.705})]{
		\includegraphics[width=0.22\textwidth]{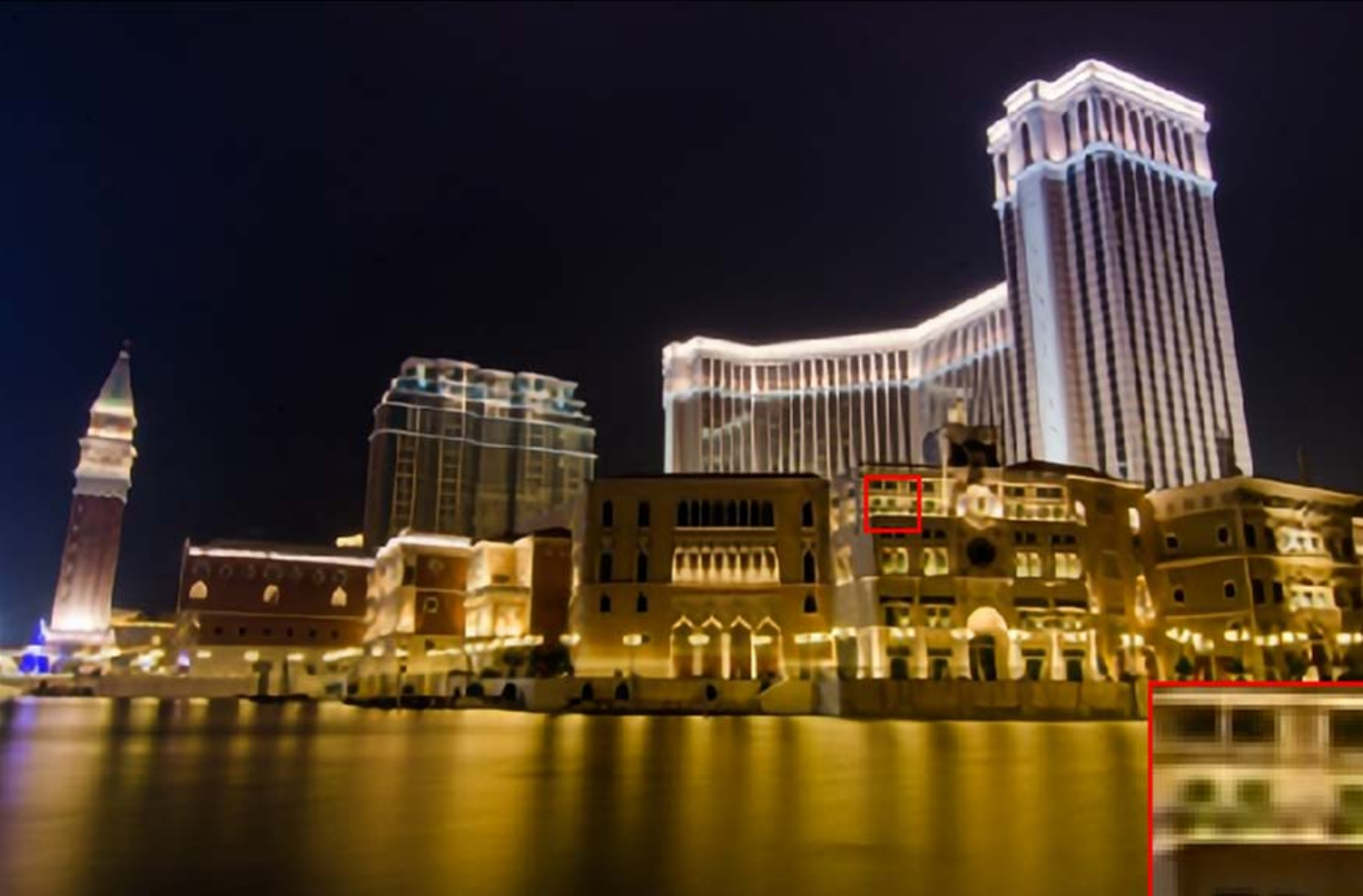}}
	\caption{The ``img085" image from the Urban100 dataset with an upscaling factor 4.}
	\label{fig:img085}
\end{figure*}

\subsection{Network analysis}
The proposed model with a global residual structure mainly learns a residual image. As show in Figure~\ref{residual_image}, the ground truth residual image mainly contains details and texture information and its normalized pixel value ranges from -0.4 to 0.5. From Figure~\ref{data_distribution}, we find that there are positive and negative values in the residual image, and the number of positive pixels is intuitively similar to that of negative ones. Obviously, the number of zero value and its neighbors is the most, which suggests that smooth region in residual image is almost eliminated. Therefore, the task of our network is to gradually subtract the smooth area of the original input image. In order to verify our intuition, we need inspect the outputs of enhancement and compression units. For better visualizing the intermediary of the proposed model, we consider an operation $T$ that can transform a $3D$ tensor $A$ to a flattened $2D$ tensor defined over the spatial dimensions, which can be formulated as follows:
\begin{equation}
T:{R^{c \times h \times w}} \to {R^{h \times w}}.
\end{equation}
Specifically, in this work, we will consider the mean of the feature maps in channel dimension, which can be described by
\begin{equation}
{T_{mean}}\left( A \right) = \frac{1}{c}\sum\nolimits_{i = 1}^c {{A_i}} ,
\end{equation}
where ${A_i} = A\left( {i,:,:} \right)$ (using Matlab notation). The average feature map can roughly represent the situations of the whole feature maps. To explore the functions of enhancement unit and compression unit, we visualize the outputs of each enhancement unit and compression unit by utilizing above-mentioned method.  As illustrated in Figure~\ref{Eunit_analyse}, from the first subpicture to the third subpicture, average feature maps gradually reduce the pixel values, especially in smooth areas. According to Figure~\ref{Eunit_analyse}, we can easily see that the first subfigure holds larger pixel values but has rough outline of the butterfly. The second and the third subfigures show that the later enhancement units continue decreasing the pixel values to obtain the features with a relatively clear contour profile. In addition, the last subfigure obviously surpasses the former figures, which brings the better inputs for the sequential compression unit that directly connects to RBlock. In summary, the function of the enhancement unit mainly enhances the outline areas of input LR image. As for the effect of compression unit, comparing Figure~\ref{Eunit_analyse} with Figure~\ref{CUnit_analyse}, we find that the pixel values of features are mapped into a smaller range through the compression unit. From the second subfigure in Figure~\ref{CUnit_analyse} and the third subfigure in Figure~\ref{Eunit_analyse}, we can see some regions of the average feature map of compression unit are enhanced by the following enhancement unit. This indicates that the process of the first three stacked blocks is to reduce the pixel value as a whole, while the last block greatly enhances the contrast between the contour and the smooth areas.

The RBlock, a transposed convolution layer, assembles the output of the final DBlock to generate the residual image. The bias term of this transposed convolution can automatically adjust the central value of the residual image data distribution to approach the ground-truth.

\subsection{Comparisons with state-of-the-arts}
We compare the proposed method with other SR methods, including bicubic, SRCNN~\cite{SRCNN,SRCNN-Ex}, VDSR~\cite{VDSR}, DRCN~\cite{DRCN}, LapSRN~\cite{LapSRN}, DRRN~\cite{DRRN} and MemNet~\cite{MemNet}. Table~\ref{tab:psnr/ssim} shows the average peak signal-to-noise ratio (PSNR) and structural similarity (SSIM) values on four benchmark datasets. The proposed method performs favorably against state-of-the-art results on most datasets. In addition, we measure all methods with information fidelity criterion (IFC) metric, which assesses the image quality based on natural scene statistics and correlates well with human perception of image super-resolution. Table~\ref{tab:ifc} shows the proposed method achieves the best performance and outperforms MemNet~\cite{MemNet} by a considerable margin.

Figure~\ref{fig:barbara}, \ref{fig:8023} and \ref{fig:img085} show visual comparisions. The ``barbara'' image has serious artifacts in the read box due to the loss of high frequency information, which can be seen from the result of bicubic interpolation. Only the proposed method recovers roughly the outline of several stacked books as shown in Figure~\ref{fig:barbara}. From Figure~\ref{fig:8023}, we can obviously see that the proposed method gains clearer contour without serious artifacts while other methods have different degrees of the fake information. In Figure~\ref{fig:img085}, the building structure on image ``img085'' of Urban100 dataset is relatively clear in the proposed method.

From Table~\ref{tab:psnr/ssim}, the performance of the proposed IDN is lower than that of MemNet in Urban100 dataset and $3 \times$, $4 \times$ scale factors, while our IDN can achieve slightly better performance in other benchmark datasets. The main reason is that MemNet takes an interpolated LR image as its input so that more information is fed into the network and the process of the SR only needs to correct the interpolated image. The algorithms that take the original LR image as input demand predicting more pixels from scratch, especially in larger images and larger magnification factors.

As for inference time, we use the public codes of the compared algorithms to evaluate the runtime on the machine with 4.2GHz Intel i7 CPU (32G RAM) and Nvidia TITAN X (Pascal) GPU (12G memory). Since we note that official implementations of MemNet and DRRN have the condition of out of the GPU memory when testing the images in BSD100 and Urban100 datasets, we divide 100 images into several parts and evaluate on these parts and then collect them for these two datasets. Table~\ref{tab:time} shows the average execution time on four benchmark datasets. It is noteworthy that the proposed IDN is approximately 500 times faster than MemNet~\cite{MemNet} with $2 \times$ magnification on the Urban100 dataset.

\section{Conclusions}
In this paper, we propose a novel network that employs distillation blocks to gradually extract abundant and efficient features for the reconstruction of HR images. The proposed approach achieves competitive results on four benchmark datasets in terms of PSNR, SSIM and IFC. Meanwhile the inference time substantially exceeds the state-of-the-art methods such as DRRN~\cite{DRRN} and MemNet~\cite{MemNet}. This compact network will be more widely applicable in practice. In the future, this approach of image super-resolution will be explored to facilitate other image restoration problems such as denosing and compression artifacts reduction.

{\small
\bibliographystyle{ieee}
\bibliography{egbib}
}

\end{document}